\documentclass[lettersize,journal]{IEEEtran}
\usepackage{amsmath,amsfonts}
\usepackage{algorithmic}
\usepackage{algorithm}
\usepackage{array}
\usepackage{textcomp}
\usepackage{stfloats}
\usepackage{url}
\usepackage{verbatim}
\usepackage{graphicx}
\usepackage[nocompress]{cite} 
\RequirePackage[section]{glossaries}
\RequirePackage[toc]{glossaries-extra} 
\RequirePackage{glossary-tree}
\makeglossaries
\setabbreviationstyle{long-short}
\setabbreviationstyle[acronym]{long-short}

\newacronym{LfD}{LfD}{learning from demonstration}
\newacronym{MCMC}{MCMC}{markov chain Monte-Carlo}
\newacronym{AI}{AI}{artificial intelligence}
\newacronym{XAI}{XAI}{explainable artificial intelligence}
\newacronym{NN}{NN}{neural network}
\newacronym{PBVS}{PBVS}{position-based visual servoing}
\newacronym{IBVS}{IBVS}{image-based visual servoing}
\newacronym{HVS}{HVS}{hybrid visual servoing}
\newacronym{CAM}{CAM}{class activation map}
\newacronym{MPC}{MPC}{model predictive control}
\newacronym{NMPC}{NMPC}{non-linear model predictive control}
\newacronym{NNMPC}{NNMPC}{neural network model predictive control}

\newacronym{PD}{PD}{proportional-derivative}
\newacronym{PID}{PID}{proportional-integral-derivative}
\newacronym{RR}{RR}{rigid robot}
\newacronym{CR}{CR}{continuum robot}
\newacronym{SMC}{SMC}{sliding mode control}
\newacronym{RL}{RL}{reinforcement learning}
\newacronym{FRL}{FRL}{fuzzy reinforcement learning}
\newacronym{CMAC}{CMAC}{cerebellar model articulation controller}
\newacronym{FNNC}{FNNC}{fuzzy neural network controller}
\newacronym{T-S}{T-S}{Takagi-Sugeno}
\newacronym{CVPC}{CVPC}{conjugated visual predictive control}
\newacronym{CV}{CV}{computer vision}
\newacronym{DOF}{DOF}{degree-of-freedom}

\newacronym{NLP}{NLP}{natural language processing}
\newacronym{DQN}{DQN}{deep Q-network}
\newacronym{PPO}{PPO}{proximal policy optimization}
\newacronym{CNN}{CNN}{convolutional neural network}
\newacronym{SOTA}{SOTA}{state-of-the-art}
\newacronym{ML}{ML}{machine learning}

\newacronym{ToF}{ToF}{time-of-flight}
\newacronym{LIDAR}{LIDAR}{light detection and ranging}
\newacronym{SLAM}{SLAM}{simultaneous localization and mapping}
\newacronym{VSLAM}{VSLAM}{visual-\gls{SLAM}}
\newacronym{VISLAM}{VISLAM}{visual-inertial-\gls{SLAM}}
\newacronym{ASLAM}{ASLAM}{active-\gls{SLAM}}

\newacronym{IMU}{IMU}{inertial measurement unit}

\newacronym{VS}{VS}{visual servoing}
\newacronym{VIO}{VIO}{visual inertial odometry}
\newacronym{VO}{VO}{visual odometry}
\newacronym{GT}{GT}{ground truth}
\newacronym{EE}{EE}{end-effector}
\newacronym{PGO}{PGO}{Pose-Graph Optimization}

\newacronym{GPS}{GPS}{global positioning system}
\newacronym{GNSS}{GNSS}{global navigation satellite system}
\newacronym{UWB}{UWB}{ultra wide-band}

\newacronym{KF}{KF}{Kalman filter}
\newacronym{EKF}{EKF}{extended Kalman filter}
\newacronym{IEKF}{IEKF}{iterative \gls{EKF}}

\newacronym{MSCKF}{MSCKF}{multi-state constraint Kalman filter}
\newacronym{SR-ISWF}{SR-ISWF}{square root inverse sliding window filter \cite{wuSquareRootInverse2015}}

\newacronym{PnP}{PnP}{perspective-n-point}
\newacronym{ICP}{ICP}{iterative closest point}
\newacronym{BA}{BA}{bundle adjustment}
\newacronym{SfM}{SfM}{structure from motion}
\newacronym{FOV}{FOV}{field of view}
\newacronym{FK}{FK}{forward kinematics}
\newacronym{IK}{IK}{inverse kinematics}

\newacronym{RGB}{RGB}{RGB}

\newacronym{SUMMIT}{SUMMIT}{SUMMIT XL from Robotnik Inc.}
\newacronym{WAM}{WAM}{WAM Arm from Barrett Inc.}
\newacronym{BHand}{BHand}{Barrett hand}
\newacronym{Summit-PC}{Summit-PC}{Adlink MXE-211}
\newacronym{WAM-PC}{WAM-PC}{Jetson Orin}
\newacronym{D455}{D455}{D455 Intel Realsense}
\newacronym{Vicon}{Vicon}{Vicon}

\newacronym{RMSE}{RMSE}{root mean squared error}
\newacronym{ATE}{ATE}{absolute trajectory error}
\newacronym{RTE}{RTE}{relative trajectory error}

\newacronym{UAV}{UAV}{unmanned aerial vehicle}
\newacronym{UGV}{UGV}{unmanned ground vehicle}
\newacronym{MHE}{MHE}{moving horizon estimator}
\newacronym{MAP}{MAP}{maximum-a-posteriori}
\newacronym{NLS}{NLS}{non-linear least squares}
\newacronym{MoE}{MoE}{mixture of experts}

\newacronym{ROS}{ROS}{Robot Operating System}
\newacronym{AVAR}{AVAR}{Allan Variance}
\newacronym{Kalibr}{Kalibr}{Kalibr toolbox}

\newacronym{AIM}{AIM}{IEEE TMECH/AIM Focused Section}
\newglossaryentry{L515}{
    name={Intel L515 camera},
    description={Intel L515 RGB-D Monocular Camera}
}
\newglossaryentry{VINS-F}{
    name={VINS-Fusion},
    description={\url{https://github.com/HKUST-Aerial-Robotics/VINS-Fusion}},
}
\newglossaryentry{VINS-M}{
    name={VINS-Mono},
    description={\url{https://github.com/HKUST-Aerial-Robotics/VINS-Mono}},
}
\newglossaryentry{VINS-Fi}{
    name={VINS-Fusion-imu},
    description={},
}
\newglossaryentry{Ceres}{
    name={Ceres},
    description={},
}
\newglossaryentry{VINS-Fg}{
    name={VINS-Fusion-gpu},
    description={},
}

\newglossaryentry{ORB}{
    name={ORB-SLAM},
    description={},
}
\newglossaryentry{ORB2}{
    name={ORB-SLAM2},
    description={},
}
\newglossaryentry{ROVIO}{
    name={ROVIO},
    description={},
}
\newglossaryentry{ALVIO}{
    name={ALVIO},
    description={},
}
\newglossaryentry{SMSCKF}{
    name={SMSCKF},
    description={},
}
\newglossaryentry{Kimera}{
    name={Kimera},
    description={},
}
\newglossaryentry{AMCL}{
    name={AMCL},
    description={},
}
\newglossaryentry{ArUco}{
    name={ArUco Markers},
    description={},
}
\newglossaryentry{KAIST}{
    name={KAIST VIO dataset},
    description={},
}
\newglossaryentry{OpenVS}{
	name={OpenVSLAM},
	description={},
}
\newglossaryentry{RTAB}{
	name={RTABMap},
	description={},
}
\newglossaryentry{RANSAC}{
	name={RANSAC},
	description={},
}

\usepackage{siunitx}

\usepackage{physics}
\usepackage{cancel}
\usepackage{amsmath,amssymb,amsfonts,environ}
\usepackage{algorithmic}
\usepackage{array}

\newcommand{\mbf}[1]{\mathbf{#1}}

\newcommand\mat[1]{\begin{matrix}#1\end{matrix}} 
\newcommand\pbracket[1]{\left( #1 \right)} 
\newcommand\bbracket[1]{\left[ #1 \right]}

\newcommand{\matb}[1]{\bbracket{ \mat{ #1 } }}

\newcommand{\hint}[1]{\textsc{#1}}

\newcommand{\set}[1]{\mathcal{#1}} 
\newcommand{\grp}[1]{\mathbb{#1}} 
\newcommand{\BS}[1]{\boldsymbol{#1}} 
\newcommand{\BN}[1]{\mathbf{#1}} 
\newcommand{\BT}[1]{\textbf{#1}} 

\newcommand{\SET}[1]{{\left\{ {#1} \right\}}}

\newcommand{\QQ}{\grp{Q}}

\newcommand{\R}{\grp{R}}

\newcommand{\SE}[1]{{\BT{SE}({#1})}}
\newcommand{\se}[1]{{\mathfrak{se}({#1})}}
\newcommand{\SO}[1]{{\BT{SO}({#1})}}

\newcommand{\zero}{\BN{0}}

\newcommand{\bv}{\BN{b}}

\newcommand{\qv}{\BN{q}}

\newcommand{\xv}{\BN{x}}

\newcommand{\pvv}{\BN{p}}
\newcommand{\vv}{\BN{v}}

\newcommand{\omegav}{\BS{\omega}}

\newcommand{\xiv}{\BS{\xi}}

\newcommand{\xspace}{\;}
\newcommand{\wrt}{\text{w.r.t.}}
\newcommand{\aka}{\text{a.k.a.}}

\newcommand{\eg}{\text{e.g.}}
\newcommand{\ie}{\text{i.e.}}

\newcommand{\I}{\mathbf{I}}

\newcommand{\HAT}[1]{\left ( {#1} \right )^\wedge}

\newcommand{\hatxi}[1]{\hat{\xiv}}
\newcommand{\hatxii}[1]{\HAT{\xiv_{#1}}}

\newcommand{\gst}{g_{st}}

\newcommand{\expxi}[1]{e^{\hatxii{#1}\theta_{#1}}}

\newcommand{\SUMMIT}{\acrshort{SUMMIT}}
\newcommand{\WAM}{\acrshort{WAM}}

\newcommand{\DOF}[1]{{#1}-\gls{DOF}\xspace }

\newcommand{\units}[1]{\left[\si{#1}\right]}

\newcommand\Tstrut{\rule{0pt}{2.6ex}}         
\newcommand\Bstrut{\rule[-0.9ex]{0pt}{0pt}}   

\newenvironment{eqconditions*}
  {\noindent\begin{tabular}{>{$}l<{$} @{${}={}$} l}}
  {\end{tabular}}
  
\newenvironment{eqparams*}
  {\par\vspace{\abovedisplayskip}\noindent\begin{tabular}{>{$}l<{$} @{${}:\text{}$} l}}
  {\end{tabular}\par\vspace{\belowdisplayskip}}

\NewEnviron{eqn}{%
    \begin{equation}
        \begin{aligned}\BODY\end{aligned}
    \end{equation} 
}

\usepackage{tikz}
\newcommand*\circled[1]{\tikz[baseline=(char.base)]{
            \node[shape=circle,draw,inner sep=0.6pt] (char) {#1};}}
            
\newcommand\REVISE[1]{#1} 
\usepackage[caption=false, font=footnotesize, labelformat=simple]{subfig}

\begin{document}

\title{A New Tightly-Coupled Dual-\acrshort{VIO} for a Mobile Manipulator with Dynamic Locomotion}
\author{
    Jianxiang Xu and Soo Jeon
    \thanks{The authors are with the Department of Mechanical and Mechatronics Engineering, University of Waterloo, Waterloo, Canada (email: jack.xu@uwaterloo.ca, soojeon@uwaterloo.ca)}
}
\markboth{IEEE/ASME TRANSACTIONS ON MECHATRONICS}%
{\MakeLowercase{\textit{et al.}}: A Draft}


\maketitle

\begin{abstract}
This paper introduces a new dual monocular visual-inertial odometry (dual-{VIO}) strategy for a mobile manipulator operating under dynamic locomotion, \ie{} coordinated movement involving both the base platform and the manipulator arm. 
Our approach has been motivated by challenges arising from inaccurate estimation due to coupled excitation when the mobile manipulator is engaged in dynamic locomotion in cluttered environments.
The technique maintains two independent monocular \acrshort{VIO} modules, with one at the mobile base and the other at the \gls{EE}, which are tightly coupled at the low level of the factor graph. 
The proposed method treats each monocular \acrshort{VIO} with respect to each other as a positional anchor through arm-kinematics. These anchor points provide a soft geometric constraint during the \acrshort{VIO} pose optimization. 
This allows us to stabilize both estimators in case of instability of one estimator in highly dynamic locomotions. 
The performance of our approach has been demonstrated through extensive experimental testing with a mobile manipulator tested in comparison to running dual \gls{VINS-M} in parallel. We envision that our method can also provide a foundation towards \gls{ASLAM} with a new perspective on multi-VIO fusion and system redundancy.
\end{abstract}

\section{Introduction \label{sec:introduction}}
\IEEEPARstart{M}{obile} robots have been increasingly adopted in warehouses, while the manipulator arm has long operated in many manufacturing plants. A mobile manipulator, a combination of both, has become an attractive candidate for various tasks requiring dynamic movement and dexterous manipulation \cite{puGeneralMobileManipulator2023a}.

Historically, the mobile robot navigation problem comprises three sub-problems; mapping, localization, and motion planning \cite{lluviaActiveMappingRobot2021a}. The first two problems are combined in the context of \gls{SLAM}, proposed by Smith et al. in 1986 \cite{chenOverviewMultiRobotCollaborative2023}. 

As the interest in this field grows, \gls{SLAM} has progressed in various forms with different sensors (\gls{IMU}, \gls{LIDAR}, \gls{ToF}, encoders and vision) with various fusion approaches (filtering or graph optimization) \cite{servieresVisualVisualInertialSLAM2021}. In particular, \gls{VSLAM} has been popular in recent years due to its low cost and rich perception \cite{chenOverviewVisualSLAM2022}. Often, \gls{VSLAM} and \gls{VO} are used interchangeably in the literature with slightly different objectives; the former focuses on the global and consistent estimate of a device's path, while the latter emphasizes the real-time pose estimations\cite{servieresVisualVisualInertialSLAM2021}. One may regard the \gls{VO} with loop-closure as a complete \gls{SLAM}. To enhance the robustness and performance of \gls{SLAM}, high rate \gls{IMU} data have been incorporated to realize \gls{VISLAM} and \gls{VIO}, which was used for a tightly-coupled fusion with vision in \gls{MSCKF}\cite{mourikisMultiStateConstraintKalman2007}.

As the \gls{SLAM} becomes more mature, a new frontier work in \glsfirst{ASLAM} has gained attention, aiming to unify all three sub-problems of navigration by bringing the task of actively planning robot locomotion to balance between exploitation (to decrease pose uncertainty) and exploration (to increase the map coverage) \cite{ahmedActiveSLAMReview2023, lluviaActiveMappingRobot2021a}. 

A mobile manipulator can be a great platform to adopt \gls{ASLAM} with additional sensor at the \glsfirst{EE} of the manipulator. However, many \gls{SLAM} modules of mobile manipulators rely solely on the base platform aided by other costly sensors such as \gls{LIDAR} \cite{heOnTheGoRobottoHumanHandovers2022}. Some use \gls{EE} localization with markers (\eg{} \gls{ArUco}) \cite{sandyConFusionSensorFusion2019, makinenRedundancyBasedVisualTool2020} which are often unusable in a rapidly changing unknown environment with dynamic locomotion \cite{sandyConFusionSensorFusion2019, puGeneralMobileManipulator2023a}. Manipulator arms provide greater spatial flexibility and precise kinematic information through high-resolution encoders. However, due to transmission gaps and mechanical vibration, the \gls{EE} may run into uncontrollable instability in global poses \cite{ubezioKalmanFilterBased2019}. As a result, the manipulator can be in an uncertain state of kinematics and dynamics \cite{luoEndEffectorPoseEstimation2021,makinenRedundancyBasedVisualTool2020}, especially in the vicinity of singularity. Consequently, the nominal calibration of \gls{IMU} noise is practically unsuitable for \gls{VIO} systems when the \gls{EE} is in dynamic motion.

Many mobile platforms use omnidirectional wheels to enable agile and holonomic movements. However, the performance of dead-reckoning estimates can be poor due to wheel slippage \cite{aguilera-marinovicGeneralDynamicModel2017, puGeneralMobileManipulator2023a}. These wheels are not ideal for \gls{VIO}, because excitations from uneven terrains can make the odometry diverge, especially when the \gls{FOV} is limited and features are poorly extracted with frequent occlusions \cite{chaeRobustAutonomousStereo2020a}. Moreover, the initialization of \gls{VIO} may become degenerative for some planar motions, resulting in inaccurate or unobservable spatiotemporal calibration \cite{caoGVINSTightlyCoupled2022,yangOnlineSelfCalibrationVisualInertial2023}. In fact, for \gls{VINS-M} \cite{qinVINSMonoRobustVersatile2018}, these motions can be problematic for its initial \gls{SfM}, causing a delayed initialization. 

Mobile manipulators are frequently equipped with an additional camera with IMU near the wrist or the EE (\aka{} the eye-in-hand vision) for perception and manipulation. This makes two VIOs readily available from the system, one at the mobile base and the other near the \gls{EE}. Combining these two sensory modules offers great flexibility when one sensor is unreliable or unavailable. However, the increased \gls{DOF} comes at a price: increased system uncertainties due to coupled excitation when engaged in dynamic movement on different unknown terrains \cite{puGeneralMobileManipulator2023a, vihonenLinearAccelerometersRate2016}. This results in highly non-uniform pose uncertainties for both \gls{VIO} modules. Current \gls{SLAM} systems often fail when the robot's motion is too complex or the environment is too challenging, \eg{} highly dynamic locomotion in cluttered environments \cite{cadenaPresentFutureSimultaneous2016}. 

\begin{figure}[!h]
    \centering
    \includegraphics[width=0.9\columnwidth]{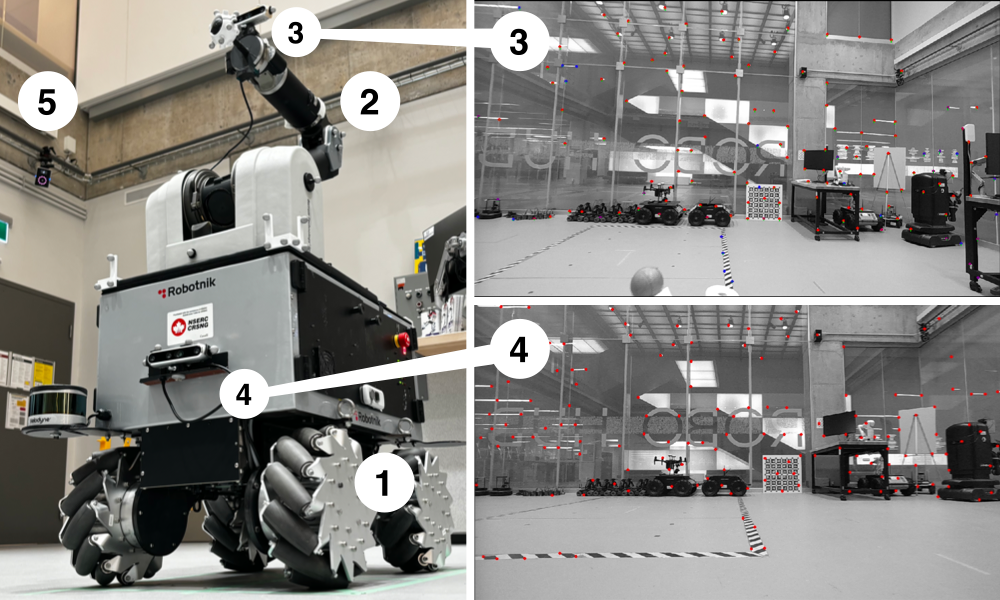}
    \caption{Our mobile manipulator comprises a \DOF{3} \SUMMIT{} mobile base \protect\circled{1} and a \DOF{7} \WAM{} robotic arm \protect\circled{2}. It has two \gls{D455} cameras configured with only monocular-\gls{RGB} and \gls{IMU}; one located at the \gls{EE} \protect\circled{3} and the other at the base \protect\circled{4}. A \gls{Vicon} system \protect\circled{5} is used for \glsfirst{GT}.}
    \label{fig:robot}
\end{figure}

This motivates us to develop a more robust low-cost real-time dual-\gls{VIO} system for a mobile manipulator with dynamic locomotion. In this paper, dynamic locomotion refers to coordinated movement involving both the base platform and the manipulator arm. Figure \ref{fig:robot} depicts a basic setup for our study, which employs one set of a monocular camera with an IMU installed on the holonomic mobile base (\DOF{3}\acrlong{SUMMIT}) and another set at the \gls{EE} of the manipulator (\DOF{7}\acrlong{WAM}). 

To the authors' best knowledge, this is the first paper introducing a tightly-coupled dynamic dual monocular \glspl{VIO} on a mobile manipulator by combining proprioception (\gls{VIO}) and kinesthesia (arm odometry) at the low level of the factor graph. Our key contributions are two-fold;

\begin{itemize}
    \item Construction of a new tightly coupled dual-\gls{VIO} approach that balances robustness and computational complexity by incorporating the arm odometry factor.
    \item Implementation of a modular distributed dual-\gls{VIO} architecture that can be applied to general configurations of mobile manipulator platforms such as the one with multiple manipulators or a legged robot with a manipulator.
\end{itemize}

\section{Related Works \label{sec:related}}
\subsection{Visual Inertial Odometry (\acrshort{VIO}) \label{sec:related-vio}}
\gls{VIO} is often divided into stereo-\gls{VIO} and monocular \gls{VIO}. We are interested in monocular \gls{VIO} due to its simplicity and low cost without losing richness in information \cite{chenOverviewVisualSLAM2022}. The stereo-\gls{VIO} allows better 3D reconstruction but requires much higher data bandwidth. Since the monocular \gls{VIO} relies on the triangulation between consecutive-frame features (as opposed to calibrated stereo triangulation), it is more prone to translational errors in the pose estimation \cite{jeonRunYourVisualInertial2021}.

Similar to most of \gls{SLAM} algorithms, \gls{VIO} consists of two main concurrent components:
\subsubsection{Front-end} The front-end module handles feature tracking and data extraction, which can be either sparse (feature) or dense (direct). The sparse methods have been predominant due to its robustness and efficiency. The performance of the front-end has a direct impact on the accuracy of the back-end \cite{strasdatVisualSLAMWhy2012}. Hence, many works focused on improving the front-end with additional features (\eg{} point and line features \cite{hePLVIOTightlyCoupledMonocular2018}, shape features\cite{aslsabbaghianhokmabadiShapedbasedTightlyCoupled2023}), or image enhancements \cite{songIRVIOIlluminationRobustVisualInertial2023}. 
\subsubsection{Back-end} The back-end module is often associated with sensor fusion. It may be categorized into either filter-based or graph-based. Filter-based methods rely on the principle of Bayes filters as their foundational form with certain relaxation  \cite{thrunProbabilisticRobotics2005}, such as Gaussian approximation with \gls{EKF} (\eg{} \gls{MSCKF} \cite{mourikisMultiStateConstraintKalman2007}). Meanwhile, most graph-based methods (\aka{} optimization-based) are derived from Bayesian inference and draw on recent advances in non-linear optimization solvers (\eg{} Ceres \cite{agarwalsameerandmierlekeirandtheceressolverteamCeresSolver2023} used in \glsfirst{VINS-M} \cite{qinVINSMonoRobustVersatile2018}). When we assume simple linear Gaussian additive noise, \gls{MAP} inference becomes equivalent to Kalman smoothing, and the tracking problem leads to the \gls{KF} \cite{dellaertFactorGraphsExploiting2021}. In practice, the filtering is preferred for high-rate and locally consistent data (\gls{IMU}), while graph-based techniques are more efficient with low-rate, globally consistent, sparse features (images).

\subsection{\gls{VIO} Fusion on Different Platforms \label{sec:diff-platform}}
Modelling errors, sensor noise, feature mis-identification, and external disturbances are inevitable, hindering accurate estimation and introducing drifts \cite{zhangRobustVisualOdometry2023}. To address these issues, \gls{VIO} is frequently integrated with other sensors (\eg{} \gls{GPS} or encoders) tailored towards a specific platform (\eg{} aerial, legged, or mobile manipulator) for different environments (\eg{} indoor or outdoor).
For instance, \glspl{UAV} often require additional fixed positioning systems to provide low-rate correction. In outdoors, \gls{GNSS} is commonly combined with \gls{VIO} through optimization (\eg{} \cite{caoGVINSTightlyCoupled2022, zhangLightweightDriftFreeFusion2023}) or filtering (\eg{} \cite{xuHighAccuracyPositioningGNSSBlocked2023} based on \gls{MSCKF}). Similarly, \gls{UWB} data can be fused with VIO (\eg{} \cite{caoVIRSLAMVisualInertial2021}) in indoors.
However, fixed positioning systems are often unavailable in cluttered and unknown environments where mobile and legged robots operate. Instead, dead reckoning sensors are used with \gls{VIO}. 
Vehicle encoders are commonly fused tightly into the \gls{VIO} through the vehicle dynamics model (\eg{} \cite{liOnlineCalibrationSingleTrack2023}) or the kinematics model (\eg{} \cite{zhuVisualInertialRGBDSLAM2022}) to improve its robustness. 

For legged robots, dynamic locomotion involves more sudden, jerky movements and impacts from ground touchdowns. These movements can significantly disrupt vision sensors, making it challenging for general \gls{VIO} to estimate the robot's state accurately \cite{zhuEventCamerabasedVisual2023}. Hence, the additional joint encoders can be tightly coupled with \gls{VIO} in the form of legged odometry factor in the factor graph to minimize the drift \cite{wisthRobustLeggedRobot2019,  kimSTEPStateEstimator2022, kangExternalForceEstimation2023}.

In mobile manipulators, similar challenges arise not only from the ground impacts but also from the dynamic coupling between the base and the arm during the dynamic locomotion. As a result, the base disturbances (from wheel-ground interactions) propagate to the arm vibration, while the arm vibrations propagate back to the base body \cite{aguilera-marinovicGeneralDynamicModel2017}. This oscillatory coupling amplifies the disturbance experienced at both the base and the \gls{EE} \gls{VIO} modules at certain modes, depending on the arm configuration.
To address this challenge, \cite{sandyConFusionSensorFusion2019} has tried to combine the \gls{VIO} at the mobile base and that at the \gls{EE} through \gls{MHE} with the help of fiducial markers. The study suggests that the \gls{IMU} is more effective in measuring the \gls{EE} when compared to the base. The results also indicate that combining the base camera with the arm odometry helps to calibrate the cameras' position with respect to the manipulator and the joint angle biases. 

These findings motivated us to develop a dual-\gls{VIO} system capable of localizing during dynamic locomotion in an unstructured, marker-free environment.  This is achieved by simultaneously maintaining two factor-graph-based \glspl{VIO} closely interacting with each other through arm kinematics. 

\subsection{Multi-\gls{VIO} Fusion \label{sec:multi-vio-fusion}}
Depending on the degree or level with which multiple sensors are fused, one may classify a general multi-sensor fusion into two categories;

\subsubsection{Loosely-coupled} The core idea of loosely-coupled fusion is to blend poses from different estimators through a weighted combination. It is a simple and efficient way to unite various sensors and odometry algorithms through a filter-based method, and thus widely used in classical multi-modal fusion (\eg{} \cite{wuMultimodalInformationFusion2022}). \cite{miillerRobustVisualInertialState2018} showed improved robustness in estimation and faster reconstruction of dense point cloud through a delayed fusion of multiple-\gls{VO} integrating \gls{IMU} into an error-state space \gls{KF}. In fact, the multi-\gls{VIO} can be considered as an example of \gls{MoE} problem \cite{morraMIXOMixtureExpertsBased2023}. However, failure of one sensor may cause deterioration or divergence of the whole estimation \cite{wangHierarchicalDistributionbasedTightlyCoupled2023}. In practice, it is not easy to characterize different uncertainties and errors \cite{morraMIXOMixtureExpertsBased2023}. 

\subsubsection{Tightly-coupled} Tightly-coupled methods prefer to fuse sensor data locally through a single optimization (\eg{}  non-overlapping multi-cameras \cite{zhangVINSMKFTightlyCoupledMultiKeyframe2018, abateMultiCameraVisualInertialSimultaneous2023, zhangBalancingBudgetFeature2022}) and solve the pose and position jointly. Often, this approach relies on the flexibility and modularity of graph optimization. It may suffer from computational complexity due to scaling issues; the network keeps growing as more optimization states are introduced \cite{morraMIXOMixtureExpertsBased2023}. 

In Multi-\gls{VIO} Fusion, all methods above improved robustness with faster reconstruction, benefiting from the increased \gls{FOV} and sensory redundancy. We propose a new technique involving a moderate tight-coupling\footnote{In some literature (\eg{} \cite{alkendiStateArtVisionBased2021}), the term "semi-tight-coupling" has been used, but we adopt "tight-coupling" as a more generally accepted term.} to balance robustness and computational complexity while maintaining the independence of each \gls{VIO} for system redundancy and failure recovery.
\subsection{Monocular-\gls{VIO} and \gls{VINS-M}} 
We are particularly interested in monocular-\gls{VIO} due to its simplicity, as discussed in Section \ref{sec:introduction}. In particular, \gls{VINS-M} has already demonstrated its robust performance with fewer resources \cite{wisthRobustLeggedRobot2019},\cite{jeonRunYourVisualInertial2021},\cite{sharafutdinovComparisonModernOpensource2023} making it the baseline of many relevant works including ours. Readers interested in \gls{VINS-F} with a monocular setup may refer to the original works \cite{qinVINSMonoRobustVersatile2018, lluviaActiveMappingRobot2021a} for in-depth fundamentals and architecture.

\section{Formulation of Proposed Dual \glspl{VIO} \label{sec:formulation}}
\begin{figure}[!h]
    \centering
    \includegraphics[width=0.9\columnwidth]{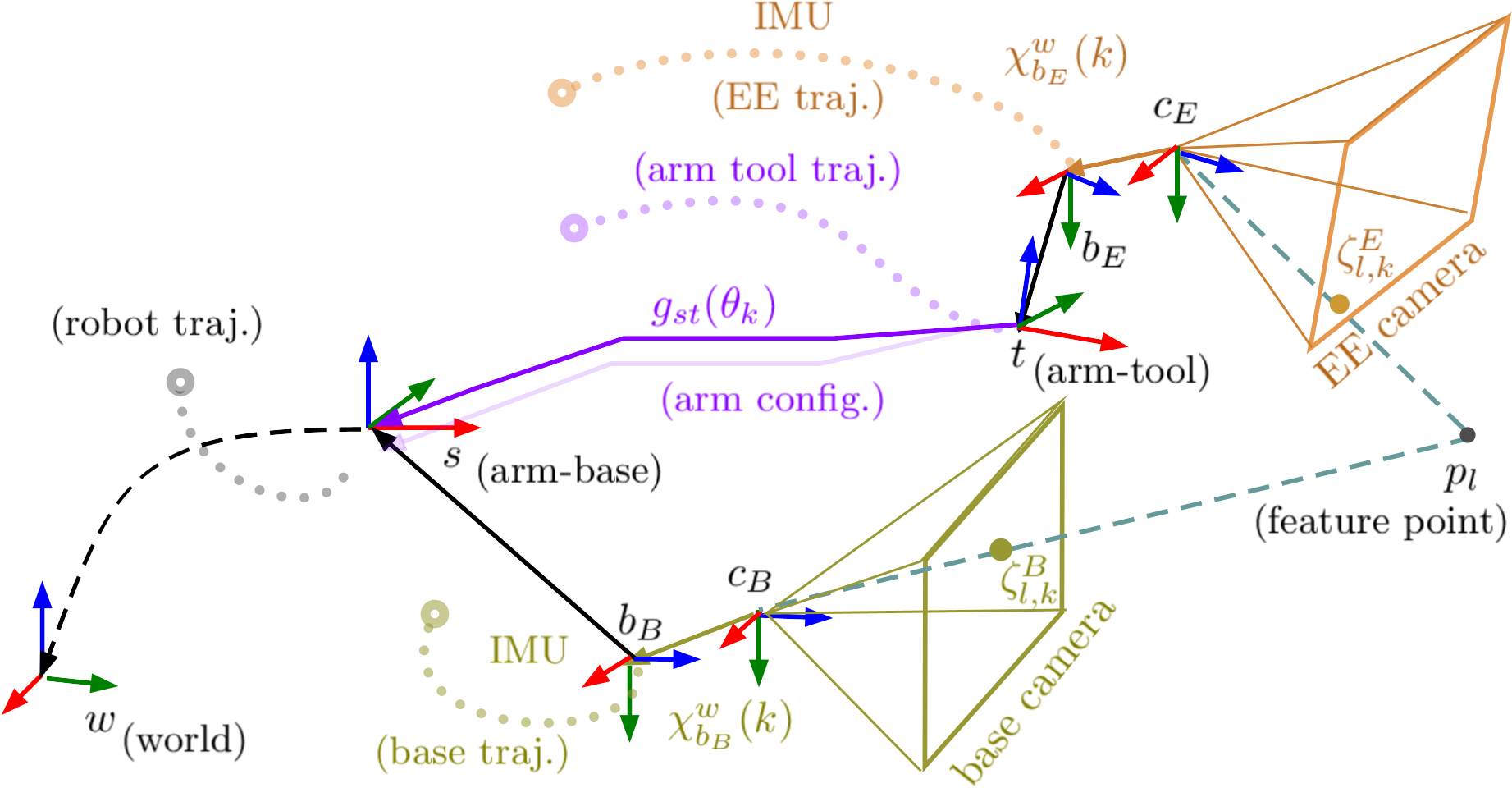}
    \caption{Frames and their relations: Solid arrows show constrained transformation; dashed arrows show unconstrained transformation; dotted lines show time-varying trajectories with a starting knot. To simplify matters, we assume that the state estimation is performed at the camera's IMU frame for both the base and the \gls{EE} (\ie{} $\chi^w_{b_B}, \chi^w_{b_E}\in\SE{3}$). The kinematic parameters for the arm are calibrated offline based on the \gls{Vicon} and 3D model at its zero-configuration, where the arm is straightened upwards ($\gst(\zero)\in\SE{3}$). \label{fig:frames}}
\end{figure}
The objective of the dual-\gls{VIO} is to estimate the poses ($\chi$\REVISE{$\in\SE{3}$}) of the camera bodies with respect to their initial poses in the world frame ($w$) both at the mobile base ($b_B$) and the \gls{EE} ($b_E$), concurrently on the fly throughout locomotion. Figure \ref{fig:frames} depicts configuration of coordinate frames used in this paper. For each motion variable (\eg{} the $\chi^w_{b_E}$), the superscript denotes its frame of reference while the subscript denotes the frame whose motion is described by the variable. The pose estimate is directly obtained from the state at frame $k$ (denoted by $\xv_{b_k}$), which is to be estimated by the factor graph (See Eq. \eqref{eqn:full-state}).
\subsection{Formulation of Factor Graph \label{sec:factor-graph}}
\begin{figure}[!h]
    \centering
    \includegraphics[width=0.9\columnwidth]{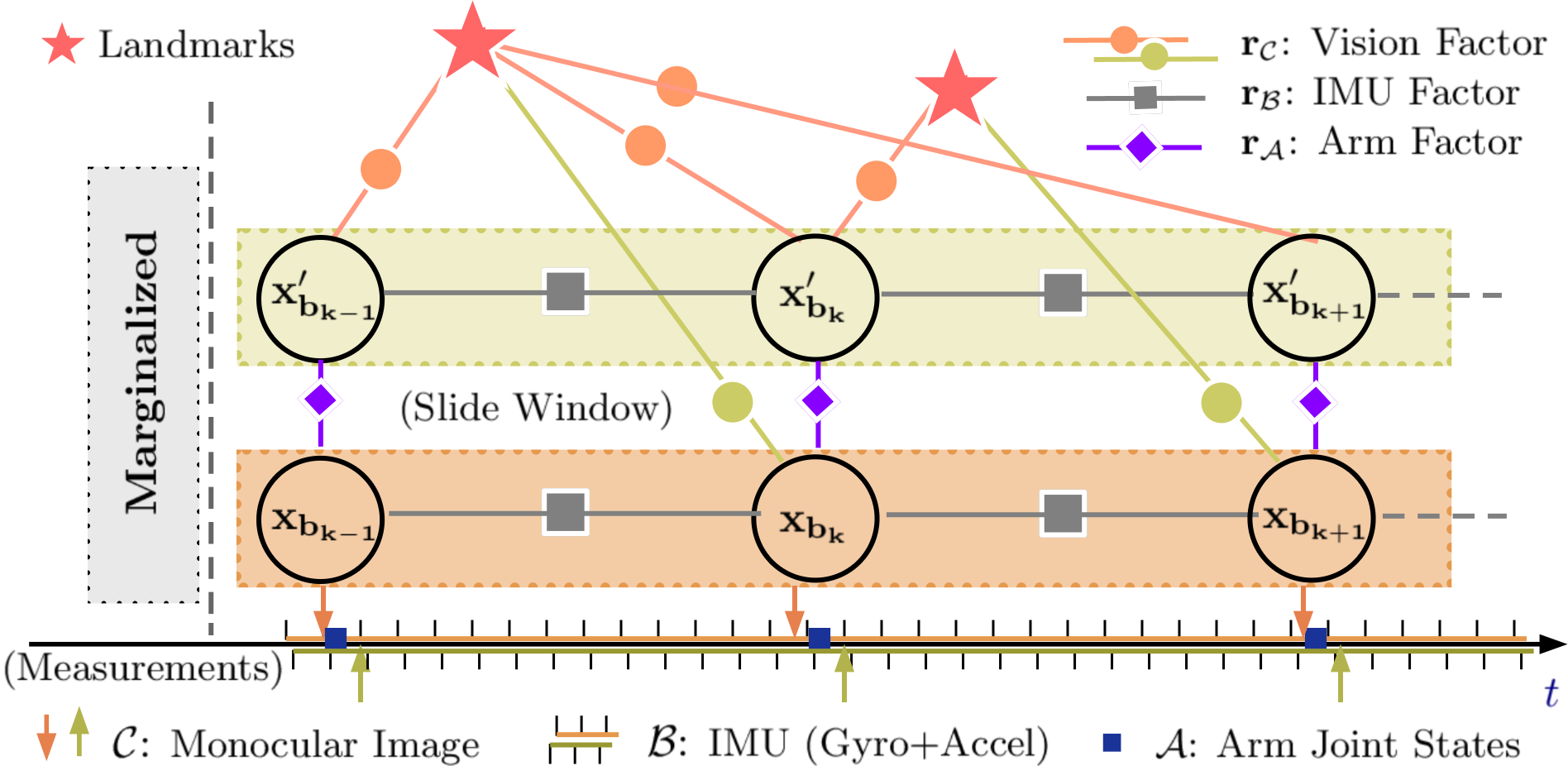}
    \caption{Illustration of the sliding window dual factor graph structure with the arm odometry factor coupling two states, with $\xv_{b_k}$ as the primary state and $\xv'_{b_k}$ as the corresponding secondary state at frame $k$. The bottom timeline shows an approximated representation of measurement timings and sampling strategy; \gls{IMU} data is sampled and pre-integrated between two frames, images are software synchronized, and arm joint states are sampled at each frame.}\label{fig:factor-graph}
\end{figure}
The proposed dual-\gls{VIO} maintains two sets of factor graphs \REVISE{running concurrently} in parallel to optimize two independent sets of states, one for the \gls{EE} and the other for the base (\ie{} $b_E$ and $b_B$, respectively, in Fig. \ref{fig:frames}). This is distinct from fusing the optimized odometry in a later stage globally (\ie{} loosely-coupled fusion) or jointly optimizing both states (\ie{} tightly-coupled fusion). Instead, a moderate coupling is introduced through the arm odometry factor from the previously optimized states at the opposing \gls{VIO}, propagated through the arm kinematics. This process is depicted in Fig. \ref{fig:factor-graph} in the form of primary and secondary \gls{VIO}. Since either \gls{VIO} is merely mirroring each other, we will denote $\xv_{b_k}$ as the primary \gls{VIO} state and \REVISE{$\xv'_{b_k}$} as the secondary one at the frame $k$. \REVISE{For instance, from the \gls{EE}'s standpoint, its primary \gls{VIO} state (i.e. $\xv_{b_k}$) is $\xv^w_{b_E}(k)$, while its complementary one ($\xv'_{b_k}$) is $\xv^w_{b_B}(k)$.} This \REVISE{relative designation} also applies to the \REVISE{geometric} pose $\chi$\REVISE{$\in\SE{3}$}, i.e. the complementary \gls{VIO} pose is $\chi'_{b_k}=\chi^w_{b_E}(k)$ if its primary VIO pose is $\chi_{b_k}=\chi^w_{b_B}(k)$.

Let us define $\set{K}:=\SET{0,..,n}$ as the set of integer indices for all the frames from $t_{k=0}$ to $t_{k=n}$ within the horizon. At each frame $k\in\set{K}$ ($k$ is also the discrete-time index for estimation within the horizon), the set of measurements from arm odometry is denoted by $\set{A}_{k}:=\SET{a_0,...,a_n}$, and that for the \gls{IMU} is denoted by $\set{Z}_{k}:=\SET{z_0,...,z_n}$ between two consecutive frames $t\in\SET{t_{k},t_{k+1}}$. Similarly, $\set{C}_{k}$ denotes the set of all the image measurements associated with features that had occurred more than once until frame $k$. We denote $\zeta_{l,j}$ as an element of $\set{C}_k$, i.e. $\zeta_{l,j}\in\set{C}_k$ ($l\in\SET{0,...,m}$, $j\in\set{K}$) to associate it explicitly with $l$-th feature in $j$-th camera frame that occurs more than once, starting from initial appearance.

Each \gls{VIO} operates over the optimization variable $\mbf{X}$ for an independent receding horizon sliding window. Thus, $\mbf{X}$ contains all state vectors $\xv_{b_k}$'s ($k\in\set{K}$) in the horizon. Besides, it also includes $\pvv^b_c$ and $\qv^b_c$ which denote the relative position and the relative orientation, respectively, between the camera frame and the corresponding body frame. For computational efficiency, $\mbf{X}$ also contains $m$ inverse feature depth (denoted by $\lambda_i$, $i\in\SET{0,1,...m}$) computed from all image measurements $\set{C}_k$. Namely, the sliding window operates its optimization over 
\begin{eqn}\label{eqn:sliding-window-variable}
\mbf{X}  = \left[\xv_{b_0}, \dots, \xv_{b_n}, \pvv^b_c, \qv^b_c, \lambda_0, \dots, \lambda_m\right] \end{eqn}
In detail, the state at each frame is a vector
\begin{eqn}\label{eqn:full-state}
    \xv_{b_k}   = \textrm{col}\left(\pvv_{b_k}^w, \vv_{b_k}^w, \qv_{b_k}^w, \bv_a, \bv_g\right), \quad k \in \set{K}
\end{eqn}
where, $\pvv_{b_k}^w \in \R^3$ denotes position, $\vv_{b_k}^w\in\R^3$ is velocity, $\qv_{b_k}^w\in\QQ\subset\R^4$ is quaternion for orientation, and $\bv_a, \bv_g \in \R^3$ denote biases of accelerometer and gyroscope, respectively. \REVISE{Note that $\chi$,  $\mbf{X}$ and $\xv$ all convey pose information while taking different mathematical forms for different purposes.}

The proposed factor graph can be solved by maximizing the posterior estimate of $\mbf{X}$ using optimization solvers (\eg{} the Ceres solver \cite{agarwalsameerandmierlekeirandtheceressolverteamCeresSolver2023}) with the objective function formulated based on \gls{BA} (similar to \cite{kangExternalForceEstimation2023, qinVINSMonoRobustVersatile2018}) as 
\begin{eqn}\label{eqn:obj-func}
    \min_{\mbf{X}}   & \SET{
        \begin{split}
            &   \norm{\mbf{r}_p - {H}_p \mbf{X}}^2 
                + 
                \sum_{k\in\set{K}} \sum_{\zeta_{l,j}\in\set{C}_k}\rho\left(\norm{\mbf{r}_\set{C}(\zeta_{l,j},\mbf{X})}^2_{P_{\zeta_{l,j}}}\right) 
            \\
             &  \REVISE{+} \sum_{k\in\set{K}}
                    \norm{\mbf{r}_{\set{Z}}({z}_{k},\mbf{X})}^2_{P_{z_{k}}} 
                + 
                \sum_{k\in\set{K}} \norm{ \REVISE{\mbf{r}_\set{A}(a_k ({\hat{\xv}'_k}), \mbf{X})} }^2_{P_{a_k}}
        \end{split}
    }   
\end{eqn}
where $\mbf{r}_p$ and ${H}_p $ in the first term come from the marginalization prior. $\mbf{r}_\set{C}$, $\mbf{r}_{\set{Z}}$ and $\mbf{r}_{\set{A}}$ denote the residuals for \REVISE{image, \gls{IMU}}, and arm odometry data, respectively. $P_{\zeta_{l,j}}, P_{z_{k}}, P_{a_k}$ are the corresponding covariance matrices used to compute the weighted information matrix for each residual (See \cite{qinVINSMonoRobustVersatile2018} for more details). The last term plays a key role in our approach by coupling two \gls{VIO}s through arm odometry (See Fig. \ref{fig:factor-graph}), which is explained in more detail below in subsection \ref{subsec:arm-odom}. 

\REVISE{Solving Eq. \eqref{eqn:obj-func} has $O(n^2)$ space complexity and $O(n^3)$ time complexity with $n$ states in the sliding window. Hence, for $M$ \gls{VIO} modules in the system, classical tightly coupled \gls{VIO}s have space and time complexity of $O(M^2n^2)$ and $O(M^3n^3)$, respectively. In comparison, the proposed method treats each \gls{VIO} as a distributed agent coupled through geometric constraints. Thus, the resulting space and time complexity would be $O(Mn^2)$ and $O(Mn^3)$, respectively.
} 

\subsection{Incorporation of Arm Odometry Factor \label{subsec:arm-odom}}
The arm odometry provides an anchor point to minimize catastrophic excitation from the current \gls{VIO} without sacrificing significant computing resources. It is computed based on the previously optimized complementary \gls{VIO} state $\SET{\hat\xv'_{b_0}, \dots \hat\xv'_{b_n}}$, where the hat notation is used to denote the solution that achieves the optimization problem \eqref{eqn:obj-func}.

\subsubsection{Manipulator Kinematics \label{sec:kinematics}}
\Gls{FK} of a manipulator maps its $N$ \gls{DOF} configuration space (which is a subset of ${\mathbb{S}^1\times \dots \times \mathbb{S}^1} \equiv \mathbb{T}^N$, \aka{} N-Torus) to the \gls{EE} pose $\chi^s_t \in \SE{3}$.
\REVISE{Using the Lie group theory for \SE{3} \cite{murrayMathematicalIntroductionRobotic1994}, we can express FK using a twist coordinate $\xiv_i \in \R^6$ for each link, which consists of the axis $\omegav_i \in \R^3$ and a point $\qv_i \in \R^3$ through it. For group operation, $\xiv_i$ can be written with the hat notation as $\HAT{\xiv_i} \in \se{3}$, i.e. Lie algebra of $\SE{3}$;
\begin{eqn}\label{eqn:twist}
    \xiv_i = \matb{- \omegav_i \times \qv_i \\ \omegav_i} = \matb{\vv_i \\ \omegav_i},
    \quad
    \hatxii{i} = \matb{\omegav_i & \vv_i \\ 0 & 0 } 
\end{eqn}
}
Accordingly, the \gls{FK} ($\gst(\BS{\theta}): \, \BS{\theta}\in\mathbb{T}^N \to \chi^s_t \in  \SE{3}$) \REVISE{in Lie group} can be defined with the product of exponential formula with each twist \REVISE{$\HAT{\xiv_i} \in \se{3}$} and corresponding angle $\theta_i \in [0,2\pi]$, \wrt{} zero configuration $\gst(\zero)$ \cite{murrayMathematicalIntroductionRobotic1994} as
\begin{eqn}\label{eqn:fk:gst}
    \gst(\BS{\theta})  = \REVISE{\expxi{1}\expxi{2}\dots\expxi{N}} \;\gst(\zero) 
\end{eqn}

\subsubsection{Arm Odometry \label{sec:arm-odometry}} At every $k$-th frame for the estimated state $\hat\xv_{b_k}$, its complementary state $\hat\xv'_{b_k}$ is firstly contracted onto the pose manifold $\hat\chi'_{b_k} \in \SE{3}$. The pose evolves through \gls{FK} $\gst(\BS{\theta})$ to the current odometry frame. The resulting estimated pose incorporating arm odometry $\tilde{\chi}_{b_k}\in \SE{3}$ is lastly lifted onto the local parameter space $a_k = (\tilde{\pvv}_{b_k}, \tilde{\qv}_{b_k}) \in \R^{3} \times \QQ$. The tilde denotes the estimated variable after incorporating arm odometry. Specifically, for the manipulator illustrated in Fig. \ref{fig:frames}, we can compute the odometry state (\ie{} the pose) over $\SE{3}$ for each of the \gls{EE} and the base \gls{VIO}'s as
\begin{eqn}
    \tilde{\chi}^w_{b_E}(k) &= \hat{\chi}^w_{b_B}(k) \; T^{b_B}_s \, \gst(\BS{\theta}) \, T^t_{b_E} \\
    \tilde{\chi}^w_{b_B}(k) &= \hat{\chi}^w_{b_E}(k) \; T^{b_E}_t \, \gst^{-1}(\BS{\theta}) \, T^s_{b_B}
\end{eqn}
where $T^{\bullet}_{\star}$ is the fixed coordinate transform between $\bullet$ and $\star$.

\subsubsection{Residual of Arm Odometry} The residual is then computed separately in translation and rotation between the current state and the arm odometry state as
\begin{eqn}
\mbf{r}_\set{A}(a_k ({\hat{\xv}'_k}), \mbf{X}) &= \matb{\delta \pvv_k \\ \delta \qv_k} =\matb{\tilde\pvv_{b_k}^{w} - \pvv_{b_k}^{w} \\ \pbracket{\qv_{b_k}^{w}}^{-1} \otimes \tilde\qv_{b_k}^{w} }
\end{eqn}
where $\otimes$ represent quaternion product \cite{baniyounesDerivationAllAttitude2019}.
\REVISE{The uncertainties at the \gls{EE} are more prone to disturbances, while the mobile platform is much more stable. Hence, the arm odometry factor imposes heavier weight for the translation residuals at the \gls{EE} (with $P^{b_E}_{a_k} = \textrm{{diag}}\left(\frac{1}{250} I_2, \frac{1}{200}, \frac{1}{10} I_4\right)$ where $I_k$ is a $k\times k$ identity matrix). Conversely, weaker factors are applied in reverse (\ie{} those at the base)  with $P^{b_B}_{a_k} = \textrm{{diag}}\left(\frac{1}{10} I_2, \frac{1}{20}, \frac{1}{10}I_4\right)$} to minimize the impact of noise, disturbance, and kinematic modeling uncertainties from the arm. \REVISE{These values for $\tiny{P_{a_k}}$ are manually tuned \wrt{} the image factor $\tiny{P_{\zeta_{l,j}}=\frac{1}{520} I_2}$}.
\section{Experiments \label{sec:experiments}}
\subsection{Experiment Setup \label{sec:experiments:setup}}
The proposed dual-\gls{VIO} is implemented based on \gls{VINS-F} under the \gls{ROS} environment. Two commercially available robotic cameras (\gls{D455} from Intel) are configured at the \gls{EE} and the mobile base as shown in Fig. \ref{fig:robot}. Each camera is enabled only with one global shutter \gls{RGB} camera (with a resolution of $1280\times720$ pixels at $30 \units{Hz}$) and one \gls{IMU} at $200 \units{Hz}$. Both depth modules are disabled due to their interference with each other and the \gls{Vicon} system. 

The \gls{IMU} noise parameters were calibrated offline through the \gls{AVAR} \cite{AllanVarianceROS2023} based on $5$-hour \gls{ROS} bag files. One collection is performed on a stationary table. Additional characteristics are collected while both cameras are on the powered robot to characterize the combined noise effects from the nominal noise in the system. The on-robot noise parameters are used for the \gls{VIO} systems.

The intrinsic parameters of each camera are calibrated offline independently with the omnidirectional camera projection model and the radial-tangential distortion model option available through \gls{Kalibr}\cite{rehderExtendingKalibrCalibrating2016} over a few minutes of the collection. The collection was performed with individual cameras moving around by hand, with the calibration board in view. Next, the camera-\gls{IMU} extrinsic parameters are calibrated with its stationary \gls{IMU} parameters.

We performed direct comparisons against the \gls{VINS-F} through the ablation study of our proposed method to evaluate the effectiveness of our approach. The baseline data is the dual-\gls{VIO} with two \gls{VINS-F} concurrently. All other hyper-parameters are kept consistent throughout the ablation study. 

\subsubsection{Environment\label{sec:exp:env}}
\begin{figure}[!h]
    \centering
    \includegraphics[width=0.85\columnwidth]{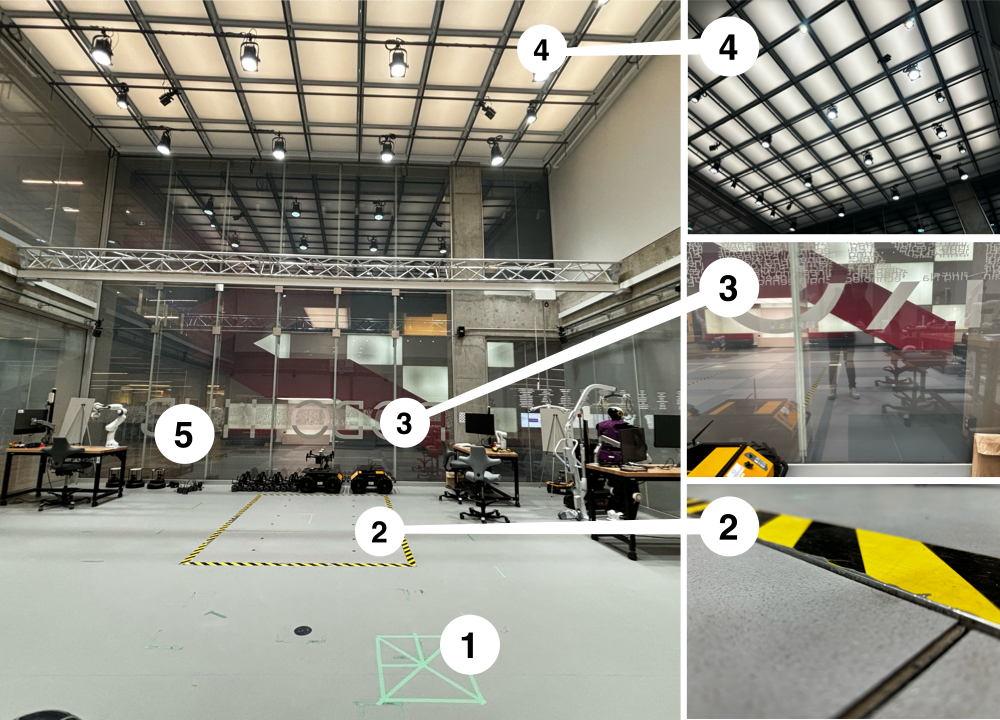}
    \caption{The environment of a cluttered facility for experiments; \protect\circled{2} uneven terrain, \protect\circled{3} reflective glass window, \protect\circled{4} low-feature illuminated ceiling, and \protect\circled{5} repetitive robots. The green \REVISE{masking tape} (\protect\circled{1}) on the floor helps \REVISE{operators locate} the robot's starting position without collision during the experiment. }
    \label{fig:env}
\end{figure}

The experiment was conducted under a well-lighted and cluttered indoor facility, which involves repetitive objects, subtle uneven terrain, low-feature ceiling, and reflective glass windows as shown in Fig. \ref{fig:env}.

\subsubsection{Choreography\label{sec:exp:dataset}}
\begin{figure}[h!]
    \centering
    \subfloat[Home (H) \label{fig:motion:arm:H}]
    {\includegraphics[width=0.28\columnwidth]{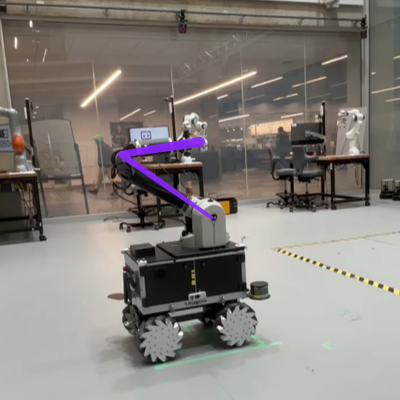}} \,
    \subfloat[Down (D) \label{fig:motion:arm:D}]
    {\includegraphics[width=0.28\columnwidth]{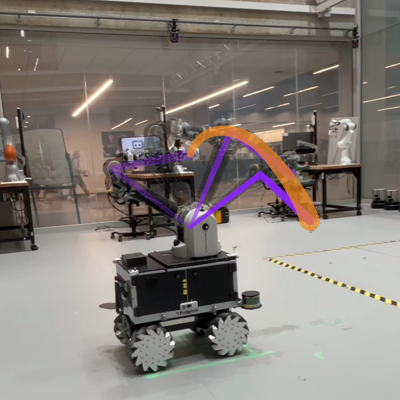}} \,
    \subfloat[Extended (E) \label{fig:motion:arm:E}]{\includegraphics[width=0.28\columnwidth]{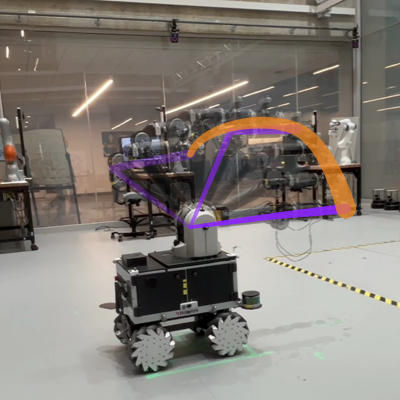}} 
    \\    
    \subfloat[Up (U) \label{fig:motion:arm:U}]
    {\includegraphics[width=0.28\columnwidth]{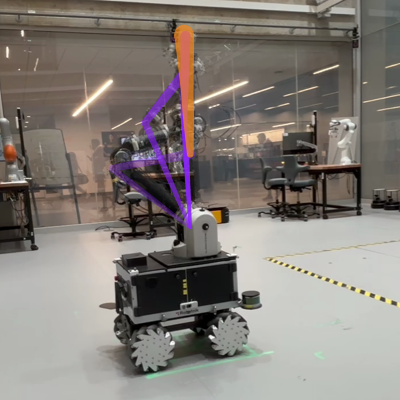}} \,
    \subfloat[Left-Right (LR) \label{fig:motion:arm:LR}]{\includegraphics[width=0.28\columnwidth]{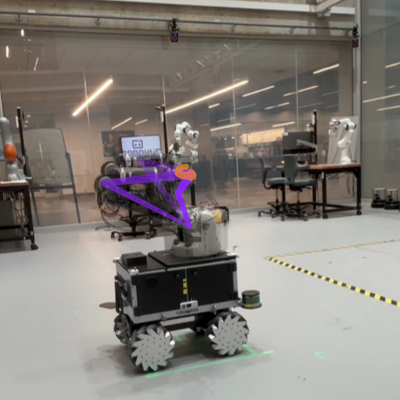}}\,
    \subfloat[Up-Down (UD) \label{fig:motion:arm:UD}]
    {\includegraphics[width=0.28\columnwidth]{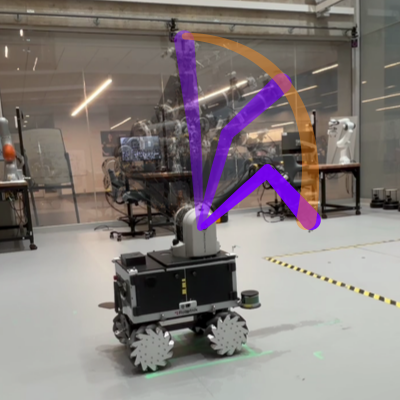}}  
    \caption{A collection of six different arm motions with base fixed. (a) showcases the home position of the arm without motion; (b), (c), and (d) demonstrate static arm motion by moving the arm to the desired configuration (\ie{} down, forward, up) initially and retracting back to the home (a) at the end of the base movements. (e) and (f) describe dynamic oscillating motions, by sweeping the arm left-and-right and up-and-down, respectively. \label{fig:motion:arm}} 
\end{figure}

The choreography of the mobile manipulator consists of 35 combinations of locomotion and arm maneuvers involving five arm motions and seven base motions. Figure \ref{fig:motion:arm} shows five different arm motions which cover both static and dynamic scenarios, starting from the home position (Fig. \ref{fig:motion:arm:H}). Figures \ref{fig:motion:arm:D}, \ref{fig:motion:arm:E}, and \ref{fig:motion:arm:U} demonstrate a simple strategy of shifting attention to less dynamic regions. Figures \ref{fig:motion:arm:LR} and \ref{fig:motion:arm:UD} explore active reconstruction to increase the mapping rate. The seven base maneuvers include three degenerative motions (spinning, forwarding, reversing) and four holonomic movements (triangle, square, circle, and bee-waggle dance). Table \ref{tab:motion:base} lists each configuration, with $v_x$ and $v_y$ denoting linear speeds, $\omega_z$ the angular rate, and $t_{\text{base}}$ and $t_{\text{EE}}$ as time duration for the base and the \gls{EE}, respectively. 
\begin{table}[!h]
    \begin{center}
    \caption{Choreography Configuration for Seven Base Movements}
    \label{tab:motion:base}
    \setlength{\tabcolsep}{3pt}
    \begin{tabular}{|lr*{7}{|c }|}
    \hline
    {} &{} & Forward & Reverse & Spin & Triangle & Square & Circle & Bee 
    \\
    {} &{} & (FWD) & (RVR) & (SPI) & (TRI) & (SQR) & (SQR) & (BEE) 
    \\ \hline
    $v_{x}$& $\tiny{\si{m/s}}$ & $0.2$  & $0.2$ & $0$ & $\pm 0.3, 0$ & $0.3, 0$ & $0.3$ & $0.3$
    \\
    $v_{y}$& $\tiny{\si{m/s}}$ & $0$  & $0$ & $0$ & $\pm 0.3, 0$ & $\pm 0.3, 0$ & $0$ & $0$   
    \\
    $\omega_{z}$& $\tiny{\si{rad/s}}$ & $0$  & $0$ & $0.2$ & $0$ & $0$ & $0.4$ & $\pm 0.4$   
    \\
    $t_{\text{base}}\hspace{-1em}$& $\tiny{\si{s}}$ & $20$ & $20$ & $30$ & $30$ & $40$ & $32$ & $50$   
    \\
    $t_{\text{EE}}$& $\tiny{\si{s}}$ & $30$ & $30$ & $40$ & $40$ & $50$ & $42$ & $60$   \\
    \hline        
    \end{tabular}
    \end{center}
\end{table}

\subsubsection{Dataset\label{sec:exp:collection}} The proposed system is evaluated through the custom dataset collected with our mobile manipulator platform in the `\verb|rosbag|' form. The dataset comprises two sets of monocular cameras and \gls{IMU} data, which were gathered through six primary arm motions (Fig. \ref{fig:motion:arm}) and seven basic base motions (Table \ref{tab:motion:base}). The dataset also includes \gls{Vicon} data, which serves as ground truth (as illustrated in Fig. \ref{fig:robot}). 

\subsubsection{Alignment\label{sec:exp:alignment}} To ensure a fair comparison between the baseline and proposed systems, the \gls{GT} datasets are offset absolutely after the initialization of each \gls{VIO} during the algorithm's runtime. This can allow us to directly compare the \gls{VIO} and corresponding \gls{GT} from the initialization point.

\subsubsection{Performance Metric\label{sec:exp:metrics}} 
Since we are interested in \gls{VIO} accuracy with dynamic locomotions, the mean of the \gls{ATE} is employed as a performance metric. We computed \gls{ATE} for translation error $(\BN{e}_{\pvv})$ and rotation error $(\BN{e}_{\qv})$ separately so that they can provide direct insights for us to understand the overall performance of the odometry. More specifically, translation error is calculated using the $l_2$ norm, while rotation error is assessed by the Frobenius norm on the $\SO{3}$ manifold towards identity as
\begin{eqn} \label{eqn:perf}
    \BN{e}_{\pvv} = \norm{\hat{\pvv}^w_{b_k} - {\pvv}^{\hint{GT}}_{b_k} }_{2} 
    \quad   
    \BN{e}_{\qv}  = \norm{\BS{R}_{\hat{\qv}^w_{b_k}}^T \, \BS{R}_{{\qv}^{\hint{GT}}_k} - \I }_{\hint{F}}
\end{eqn}
where ${\pvv}^{\hint{GT}}_{b_k}$ and $\BS{R}_{{\qv}^{\hint{GT}}_k}$ denote the position vector and the rotation matrix, respectively, of the \gls{GT} data, and $\BS{R}_{\hat{\qv}^w_{b_k}}$ denotes the rotation matrix for the estimated quaternion $\hat{\qv}^w_{b_k}$. \REVISE{For more details of these metrics and others, readers may refer to  \cite{zhangTutorialQuantitativeTrajectory2018}.}

\begin{figure*}[t!]
    \centering
    \subfloat[Base translation \acrshort{ATE} metrics with up-down arm motion (UD) \label{fig:result:UD:base:P}]{\includegraphics[width=\columnwidth]{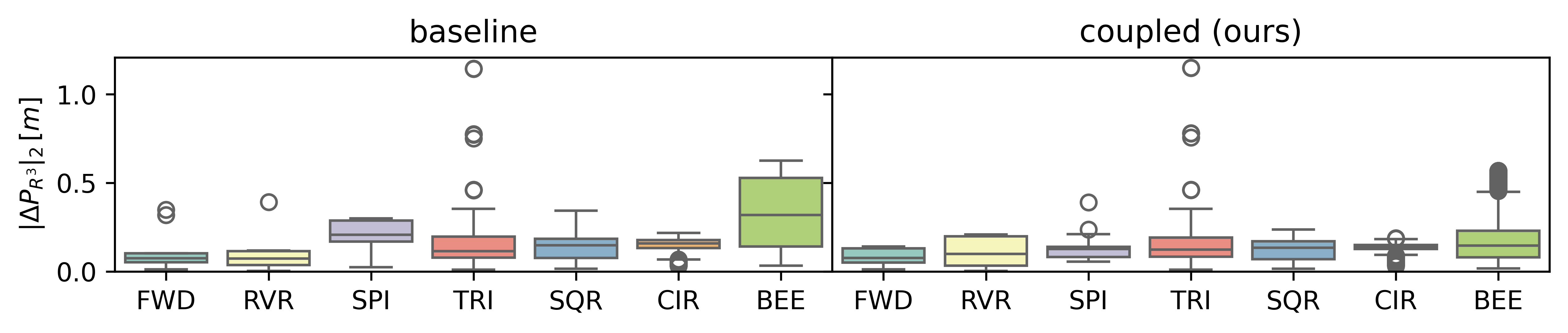}}
    \;
    \subfloat[Base orientation \acrshort{ATE} metrics with up-down arm motion (UD) \label{fig:result:UD:base:R}]{\includegraphics[width=\columnwidth]{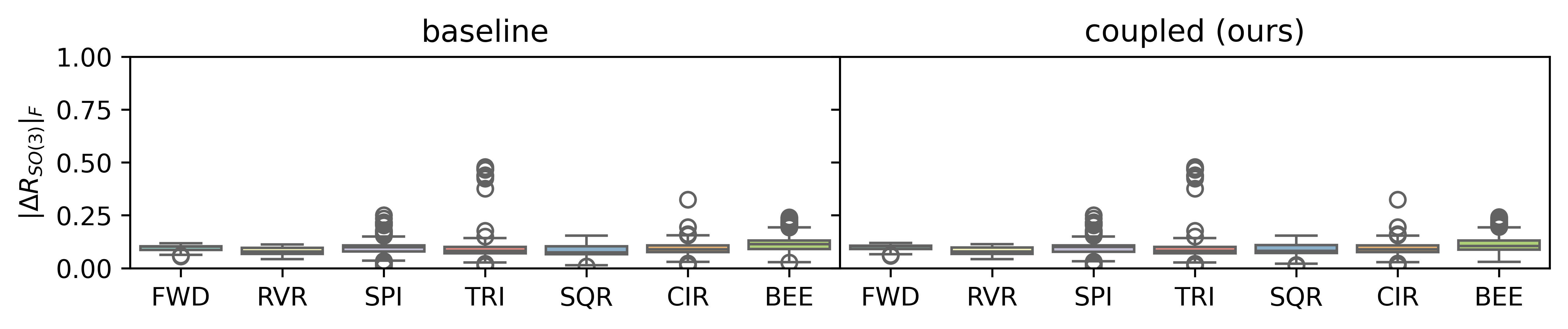}}
    \\ \vspace{-1em} 
    \subfloat[\gls{EE} translation \acrshort{ATE} metrics with up-down arm motion (UD) \label{fig:result:UD:EE:P}]{\includegraphics[width=\columnwidth]{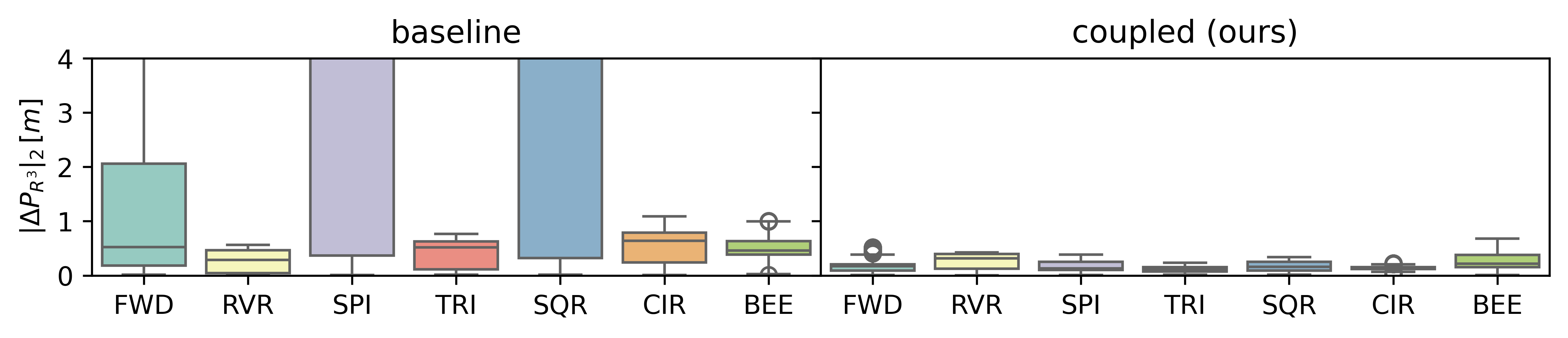}}
    \;
    \subfloat[\gls{EE} orientation \acrshort{ATE} metrics with up-down arm motion (UD) \label{fig:result:UD:EE:R}]{\includegraphics[width=\columnwidth]{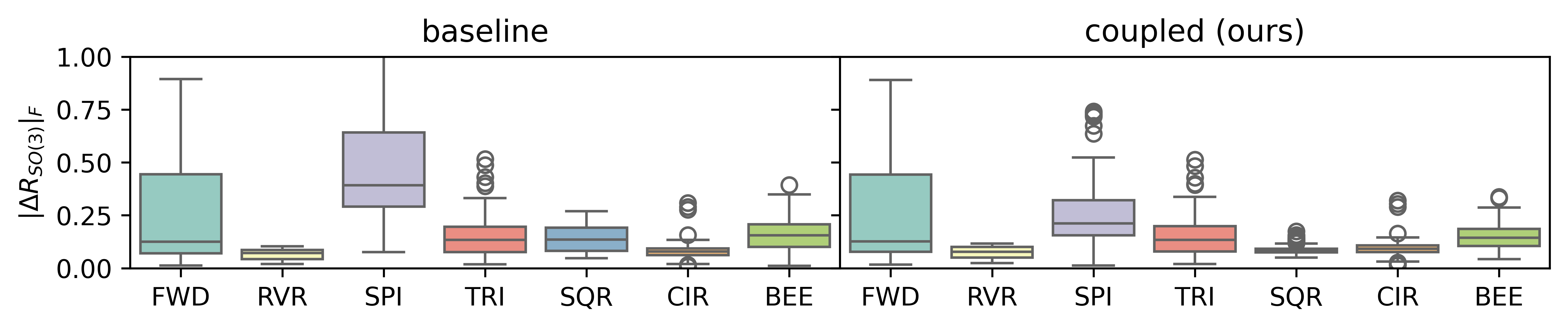}}
    \\ \vspace{-1em} 
    \subfloat[FWD Base \label{fig:result:UD:fwd}]{\includegraphics[width=0.29\columnwidth]{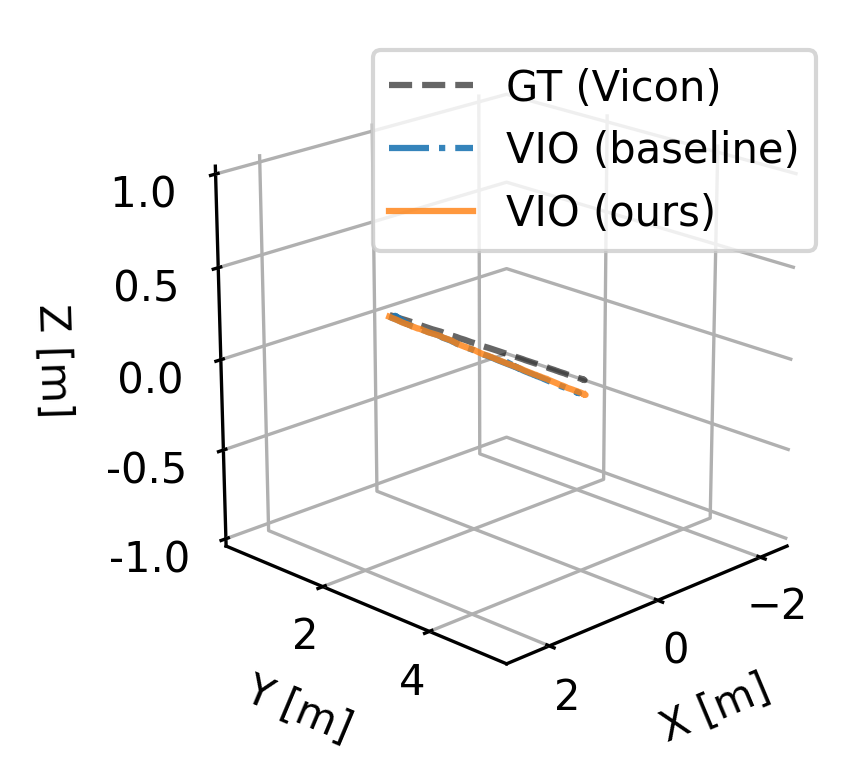}}
    \subfloat[RVR Base \label{fig:result:UD:rvr}]{\includegraphics[width=0.29\columnwidth]{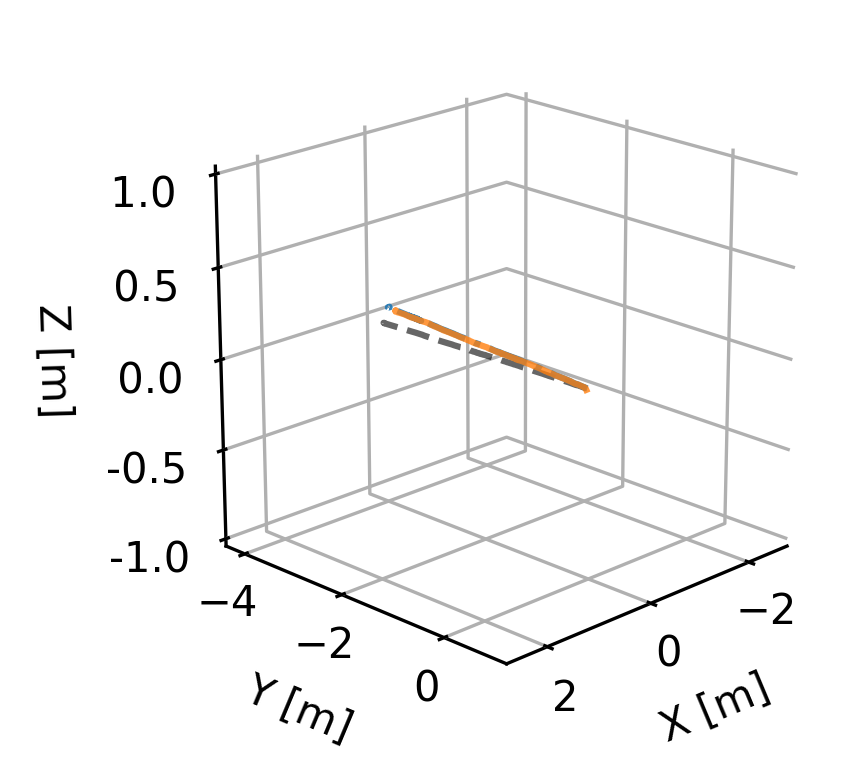}}
    \subfloat[SPI Base \label{fig:result:UD:spi}]{\includegraphics[width=0.29\columnwidth]{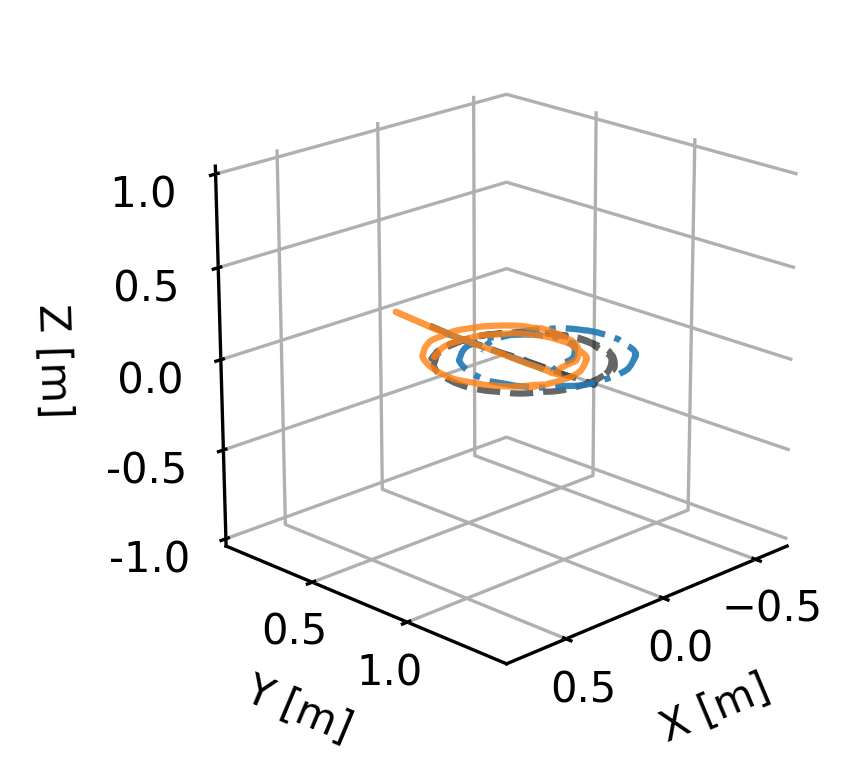}}
    \subfloat[TRI Base \label{fig:result:UD:tri}]{\includegraphics[width=0.29\columnwidth]{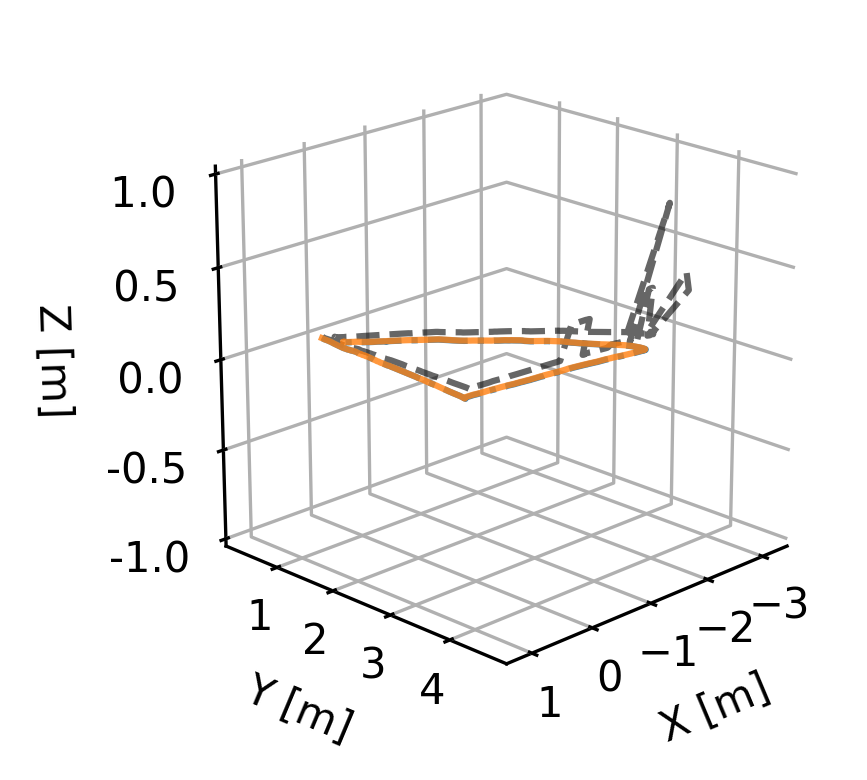}}
    \subfloat[SQR Base \label{fig:result:UD:sqr}]{\includegraphics[width=0.29\columnwidth]{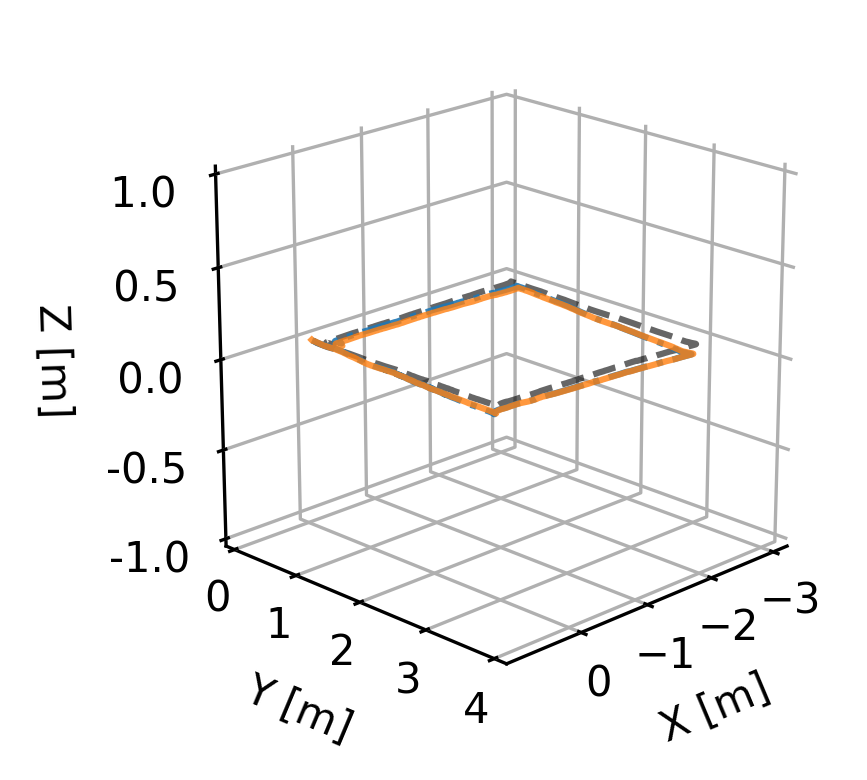}}
    \subfloat[CIR Base \label{fig:result:UD:cir}]{\includegraphics[width=0.29\columnwidth]{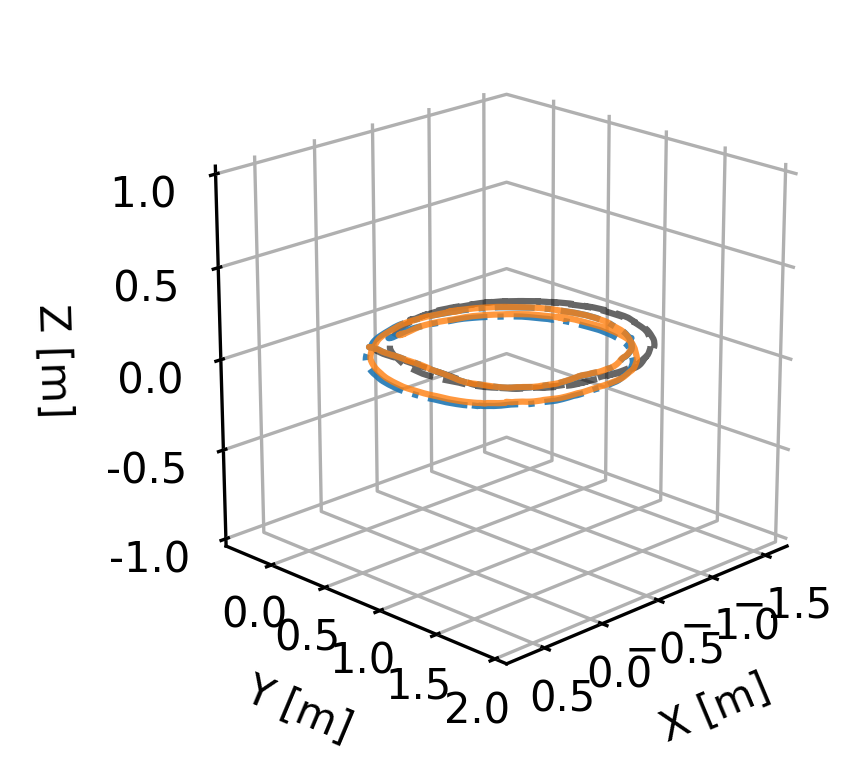}}
    \subfloat[BEE Base \label{fig:result:UD:bee}]{\includegraphics[width=0.29\columnwidth]{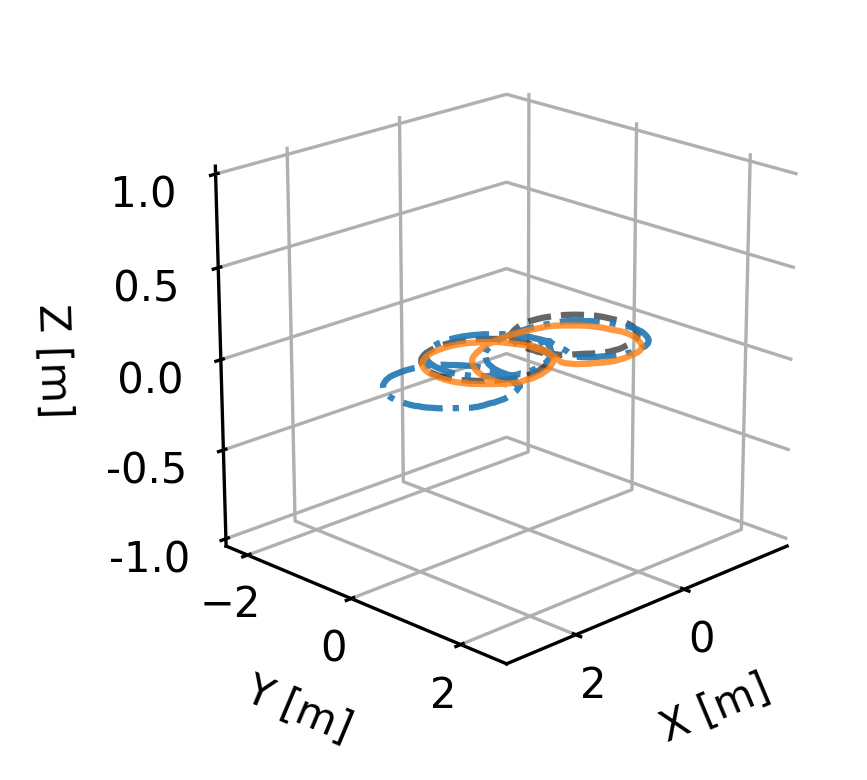}}
    \\ \vspace{-1em} 
    \subfloat[FWD UD EE \label{fig:result:UD:fwd}]{\includegraphics[width=0.29\columnwidth]{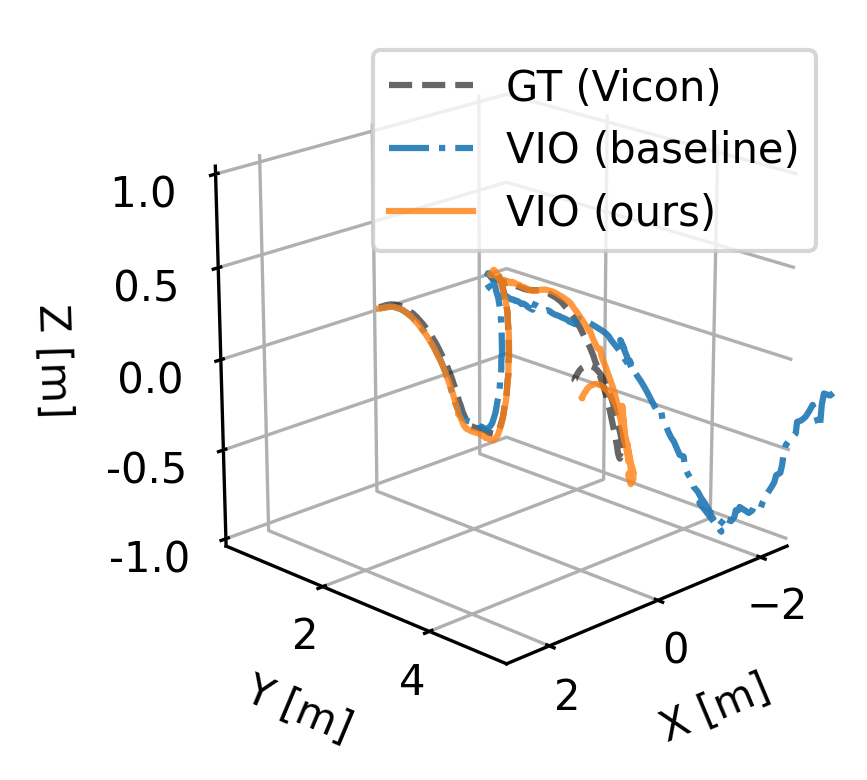}}
    \subfloat[RVR UD EE \label{fig:result:UD:rvr}]{\includegraphics[width=0.29\columnwidth]{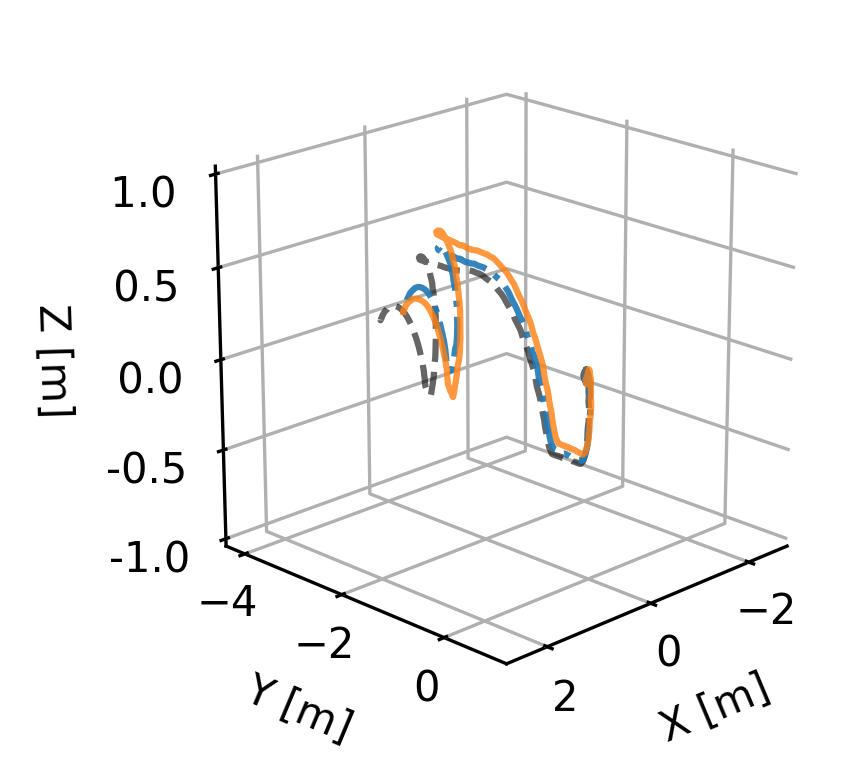}}
    \subfloat[SPI UD EE \label{fig:result:UD:spi}]{\includegraphics[width=0.29\columnwidth]{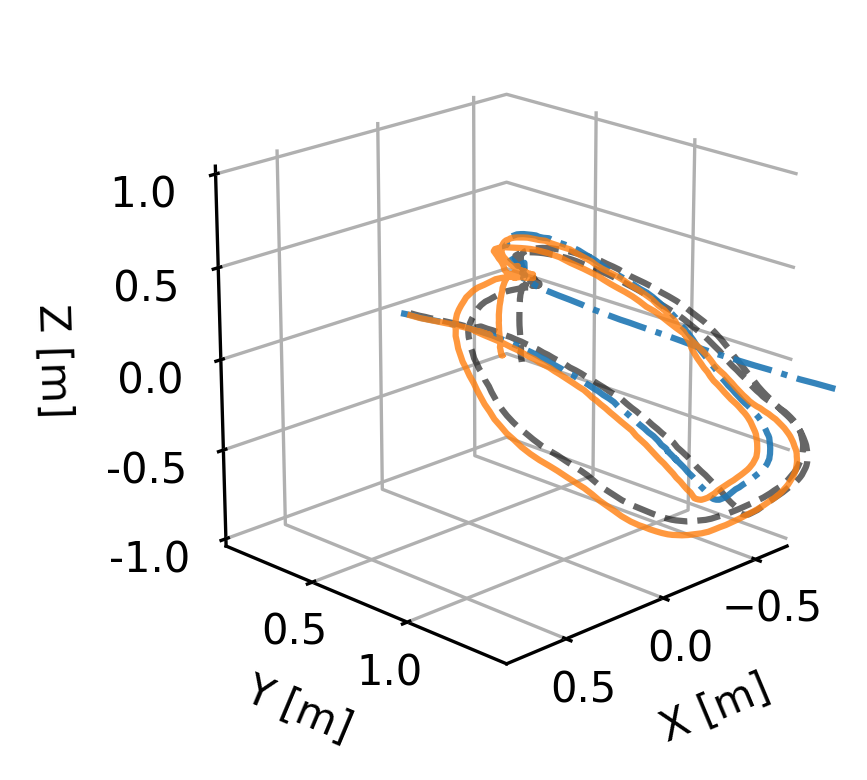}}
    \subfloat[TRI UD EE \label{fig:result:UD:tri}]{\includegraphics[width=0.29\columnwidth]{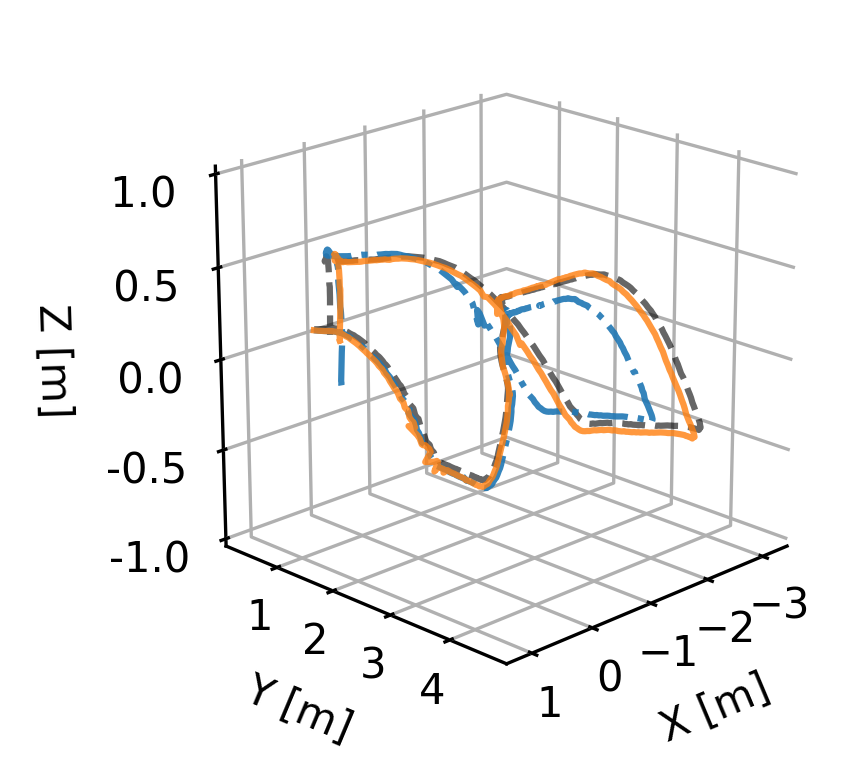}}
    \subfloat[SQR UD EE \label{fig:result:UD:sqr}]{\includegraphics[width=0.29\columnwidth]{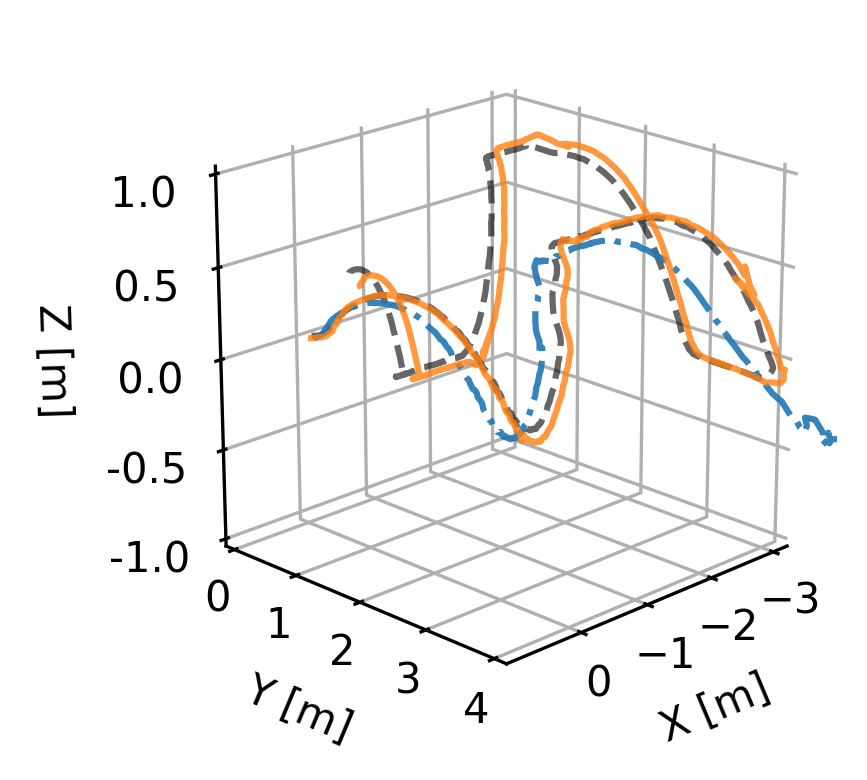}}
    \subfloat[CIR UD EE \label{fig:result:UD:cir}]{\includegraphics[width=0.29\columnwidth]{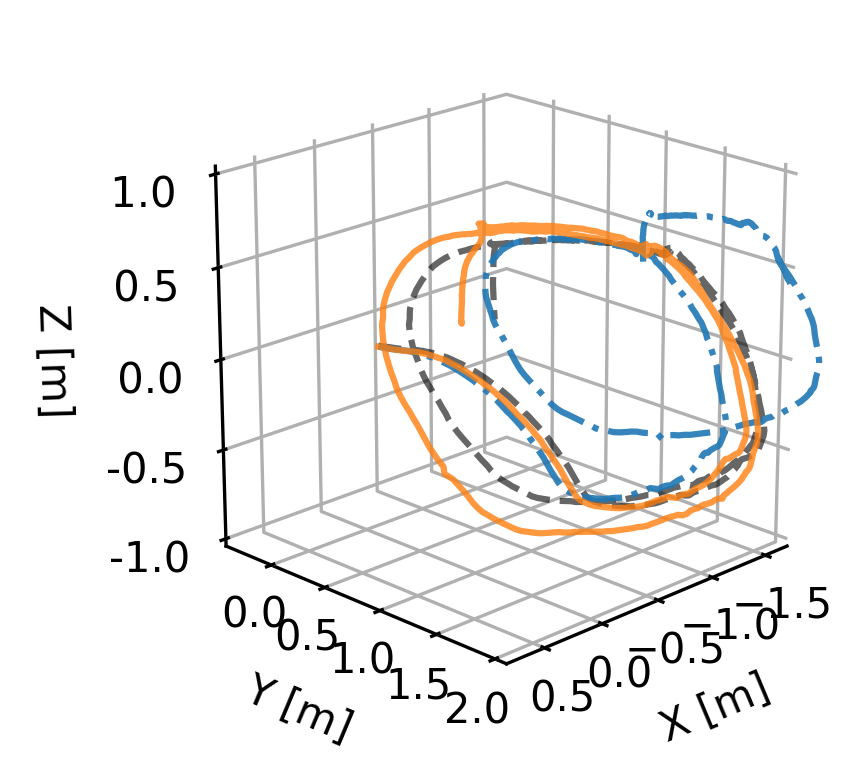}}
    \subfloat[BEE UD EE \label{fig:result:UD:bee}]{\includegraphics[width=0.29\columnwidth]{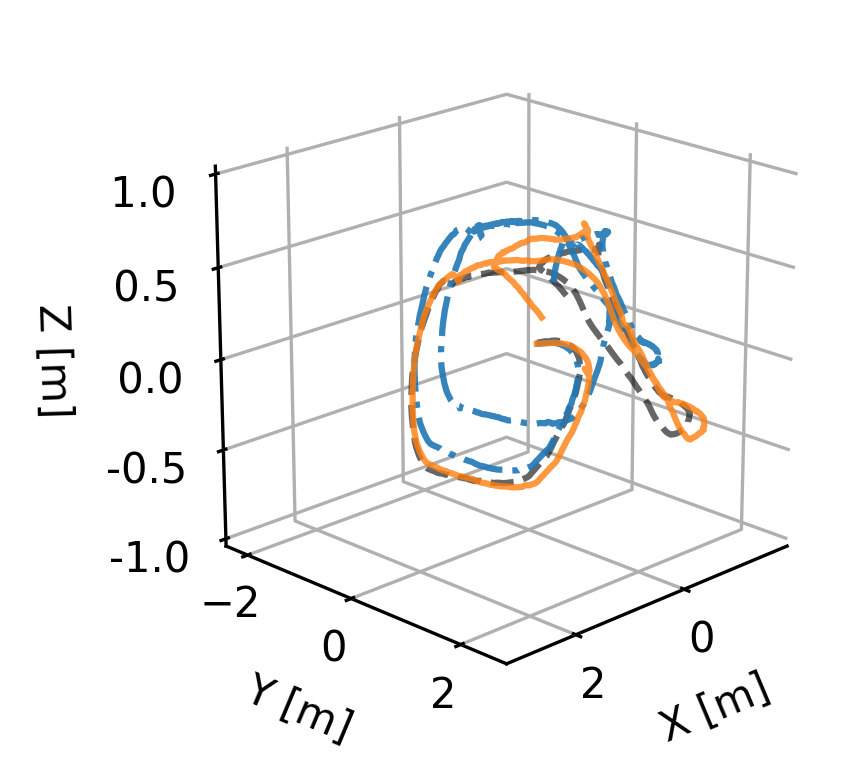}}
        \\ \vspace{-0.5em} 
    \subfloat[FWD UD Video \label{fig:result:UD:fwd:vid}]{\includegraphics[height=34pt,trim={0 5cm 0 5cm},clip]{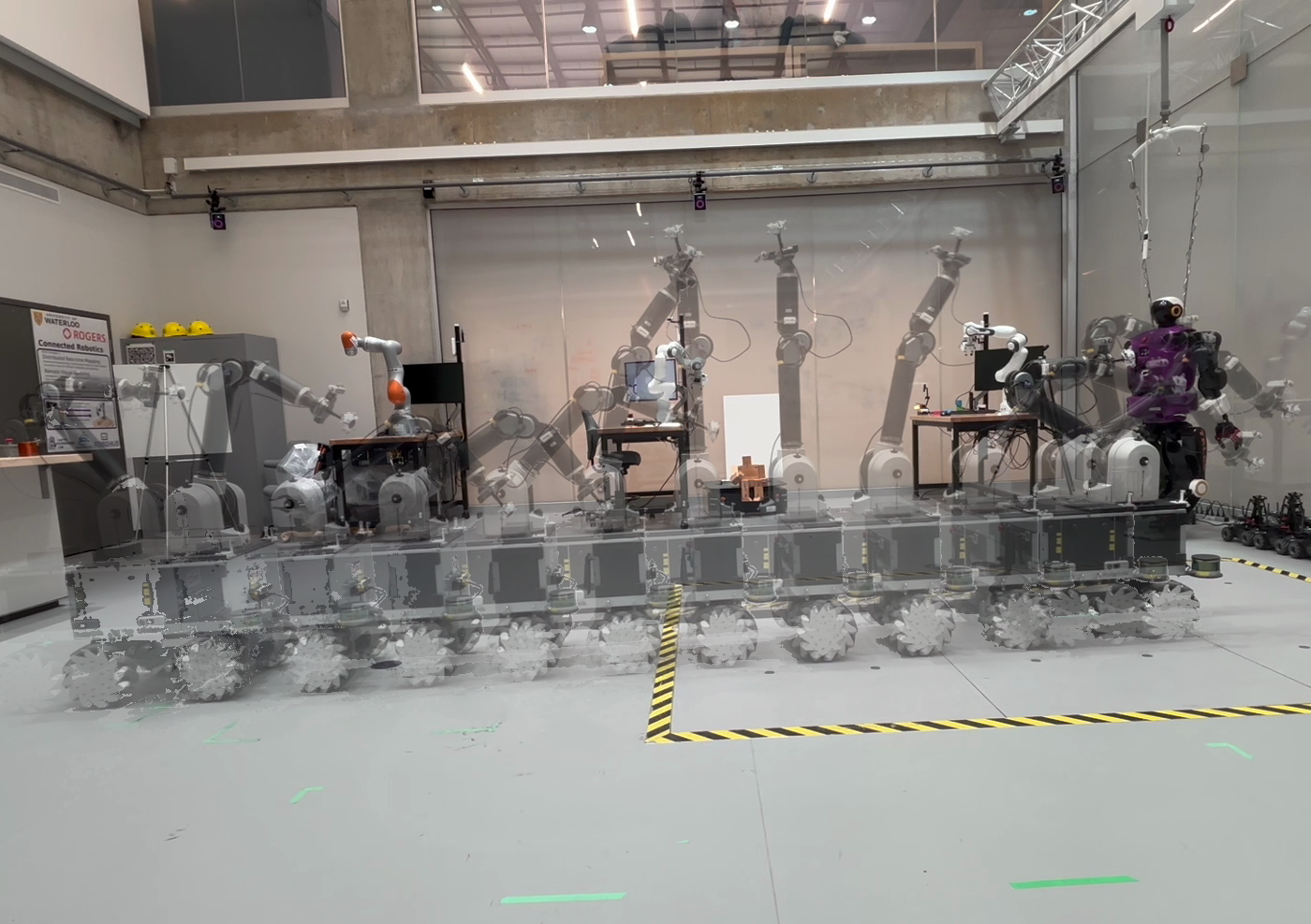}}\,
    \subfloat[RVR UD Video \label{fig:result:UD:rvr:vid}]{\includegraphics[height=34pt,trim={0 5cm 0 5cm},clip]{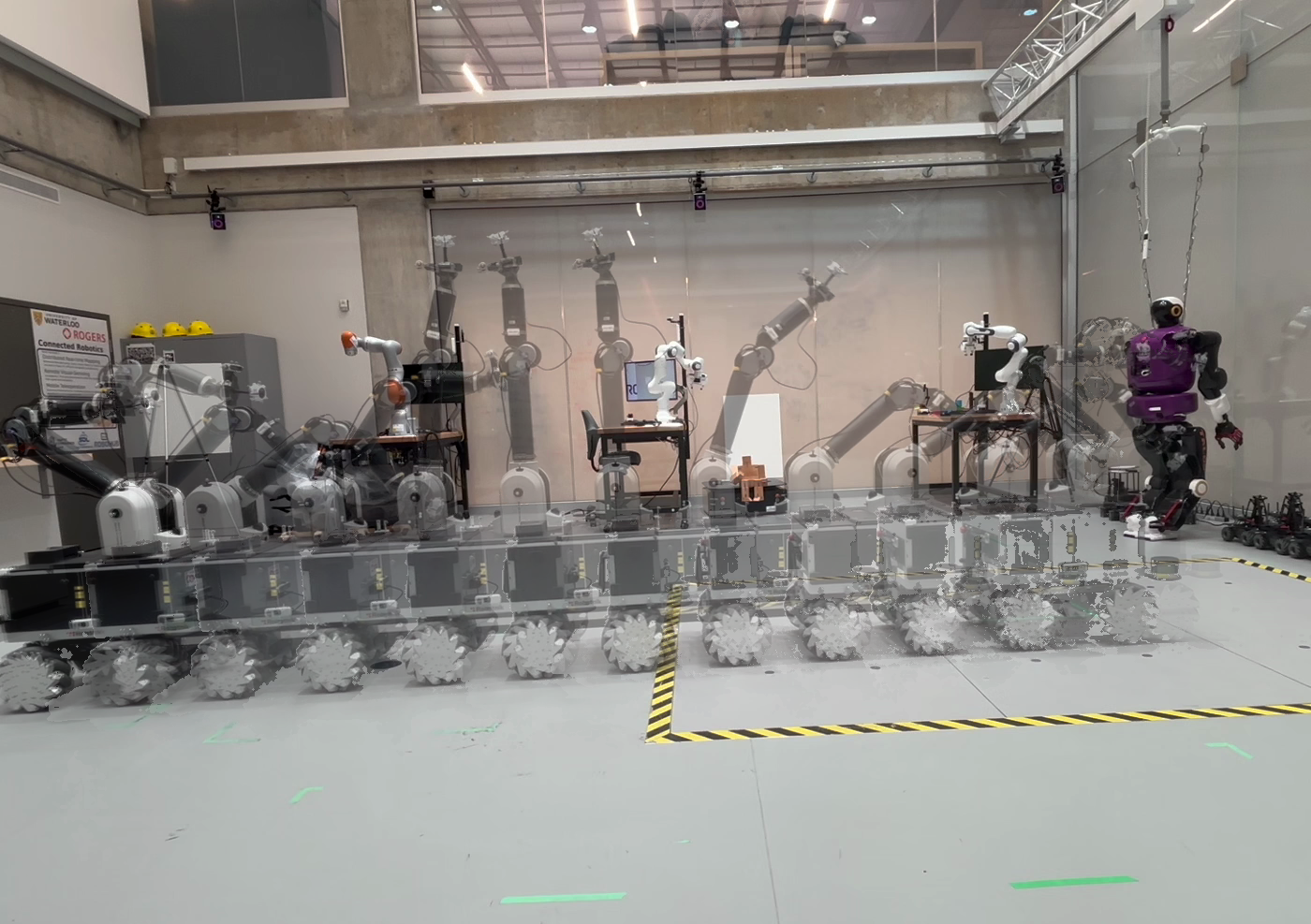}}\,
    \subfloat[SPI UD Video \label{fig:result:UD:spi:vid}]{\includegraphics[height=34pt,trim={0 5cm 0 5cm},clip]{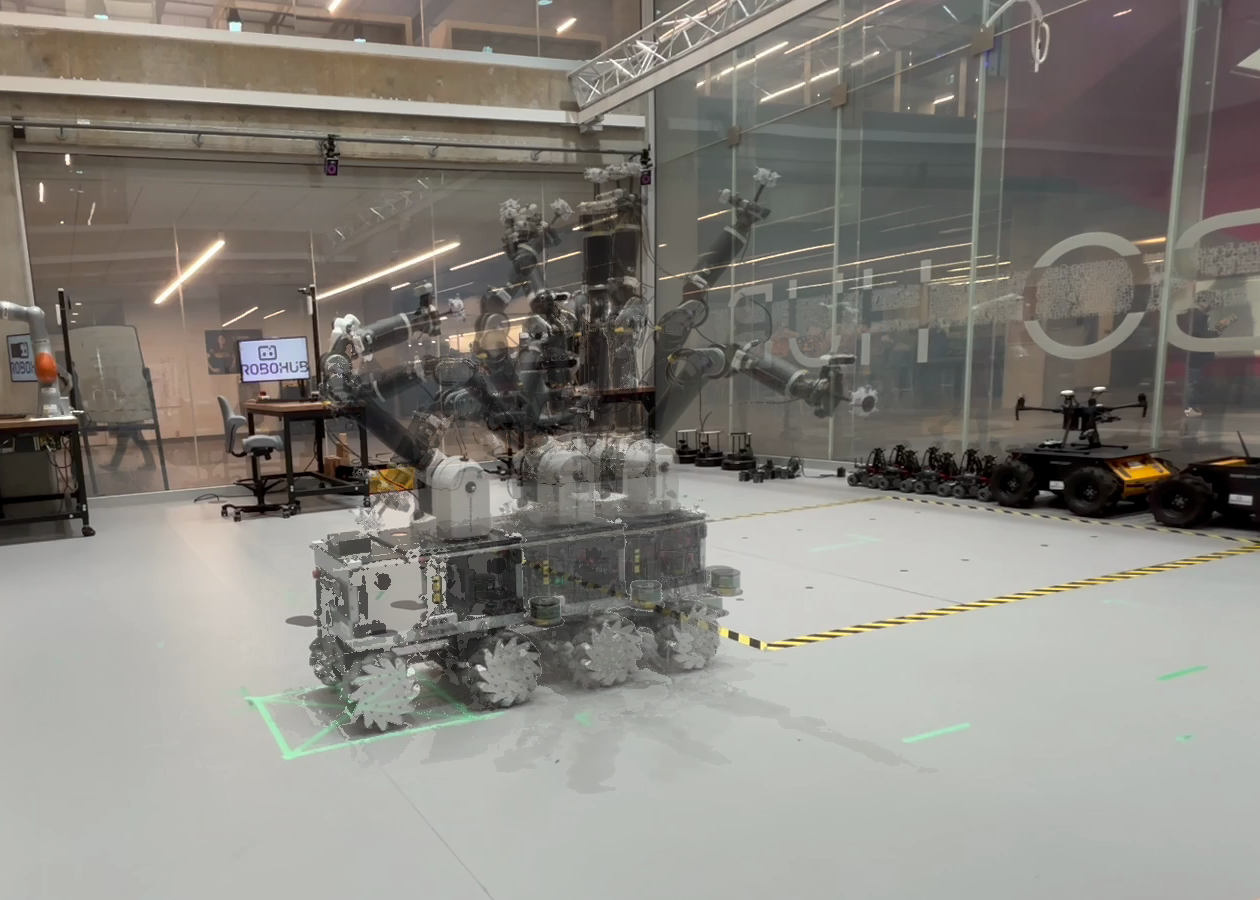}}\,
    \subfloat[TRI UD Video \label{fig:result:UD:tri:vid}]{\includegraphics[height=34pt,trim={0 5cm 0 5cm},clip]{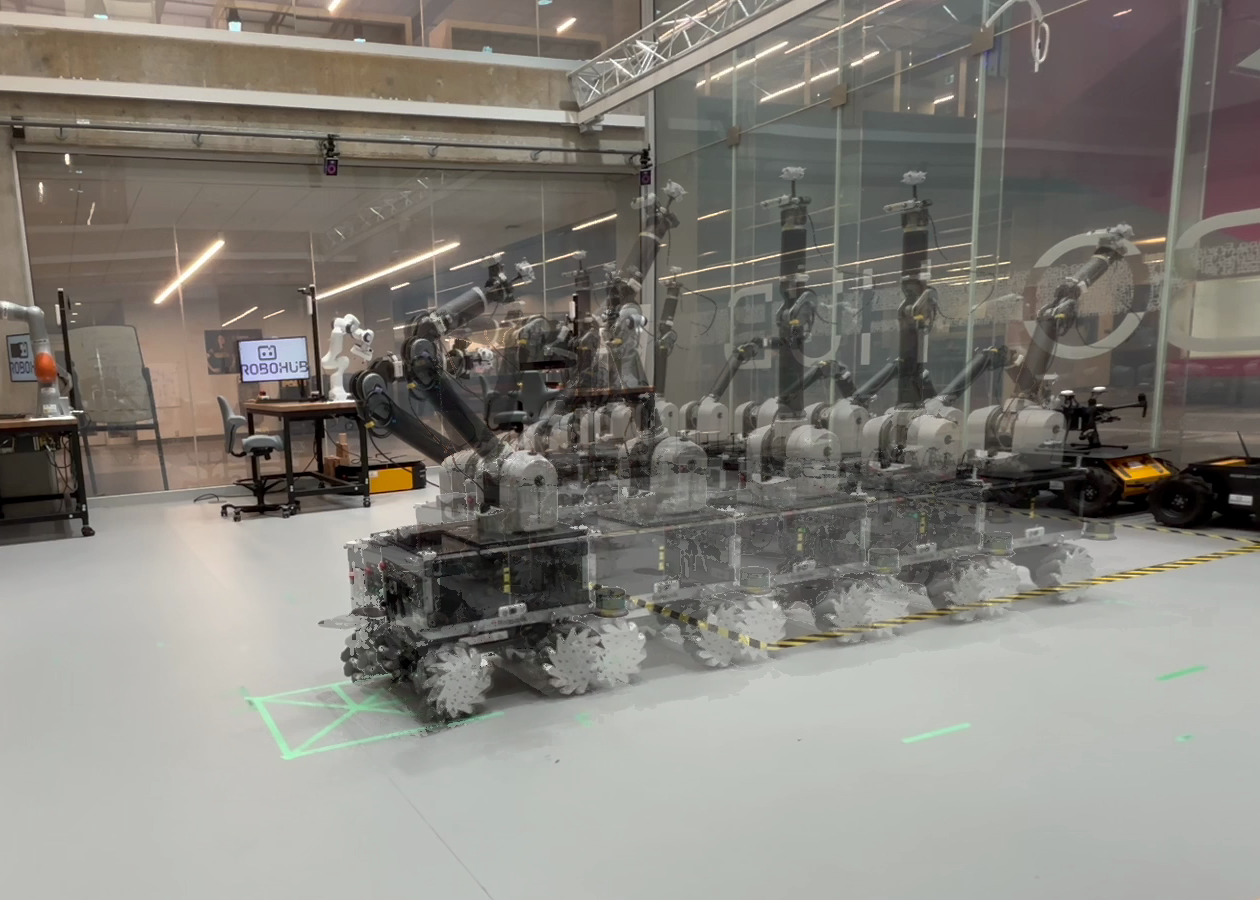}}\,
    \subfloat[SQR UD Video \label{fig:result:UD:sqr:vid}]{\includegraphics[height=34pt,trim={0 5cm 0 5cm},clip]{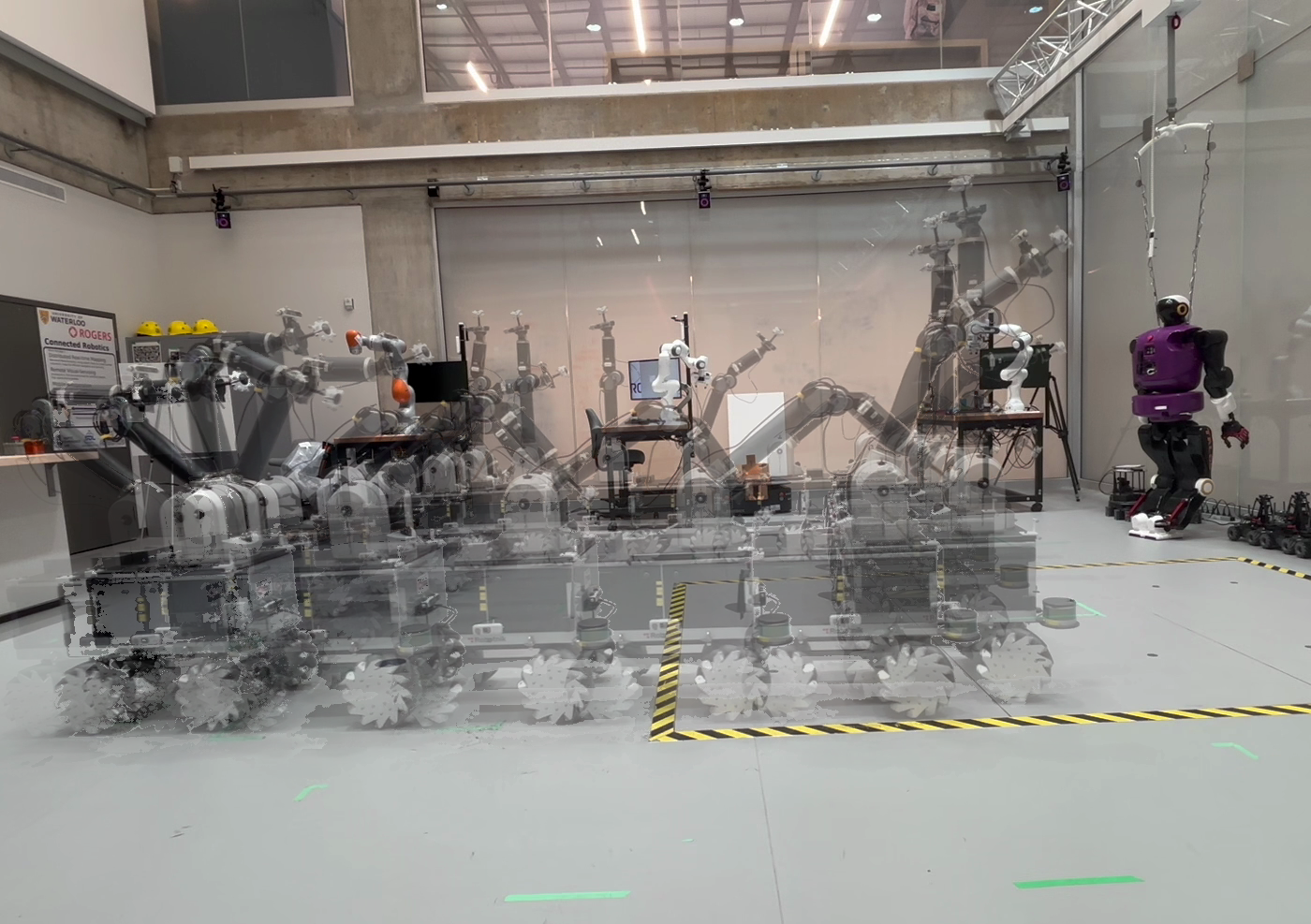}}\,
    \subfloat[CIR UD Video \label{fig:result:UD:cir:vid}]{\includegraphics[height=34pt,trim={0 5cm 0 5cm},clip]{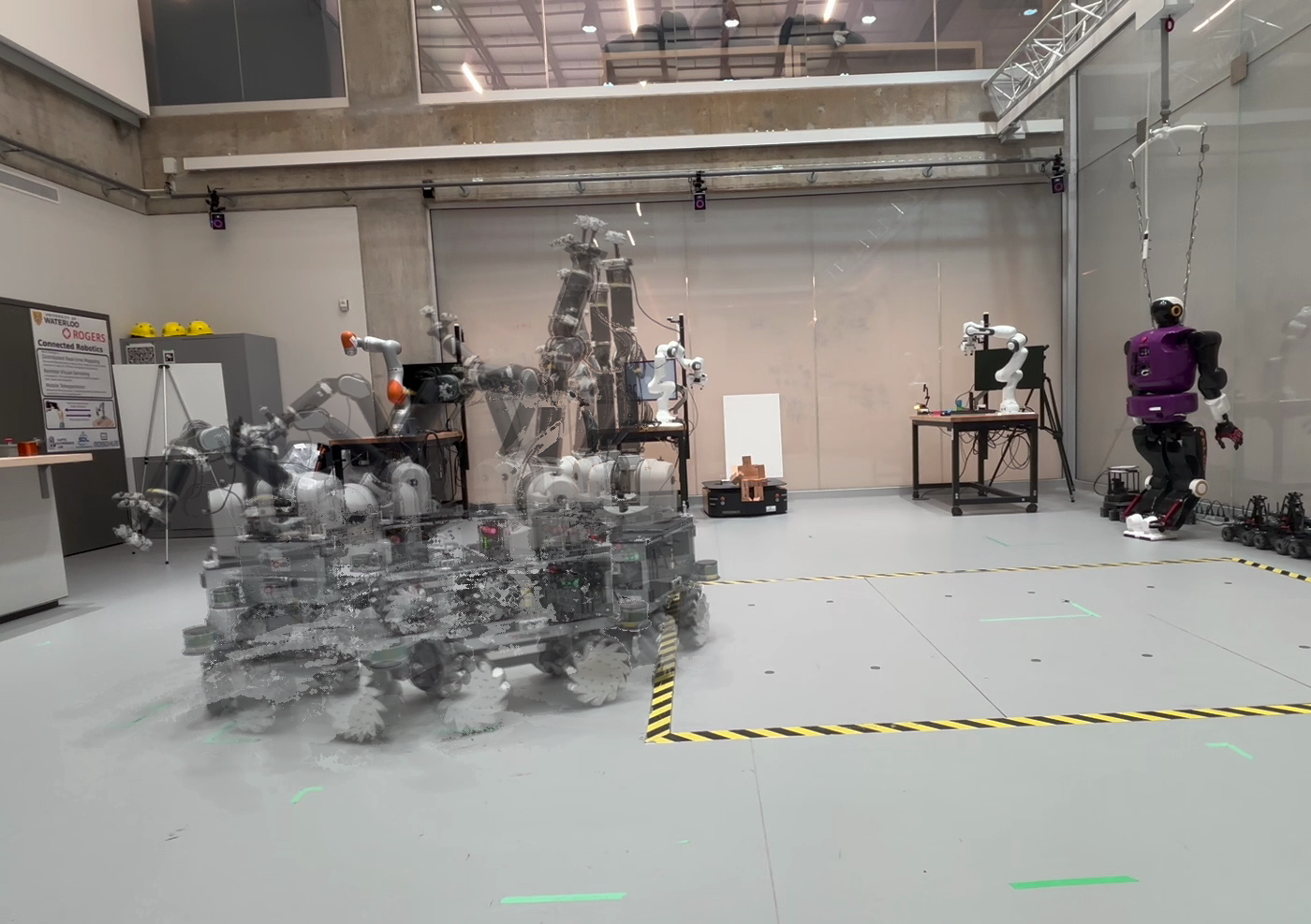}}\,
    \subfloat[BEE UD Video \label{fig:result:UD:bee:vid}]{\includegraphics[height=34pt,trim={0 5cm 0 5cm},clip]{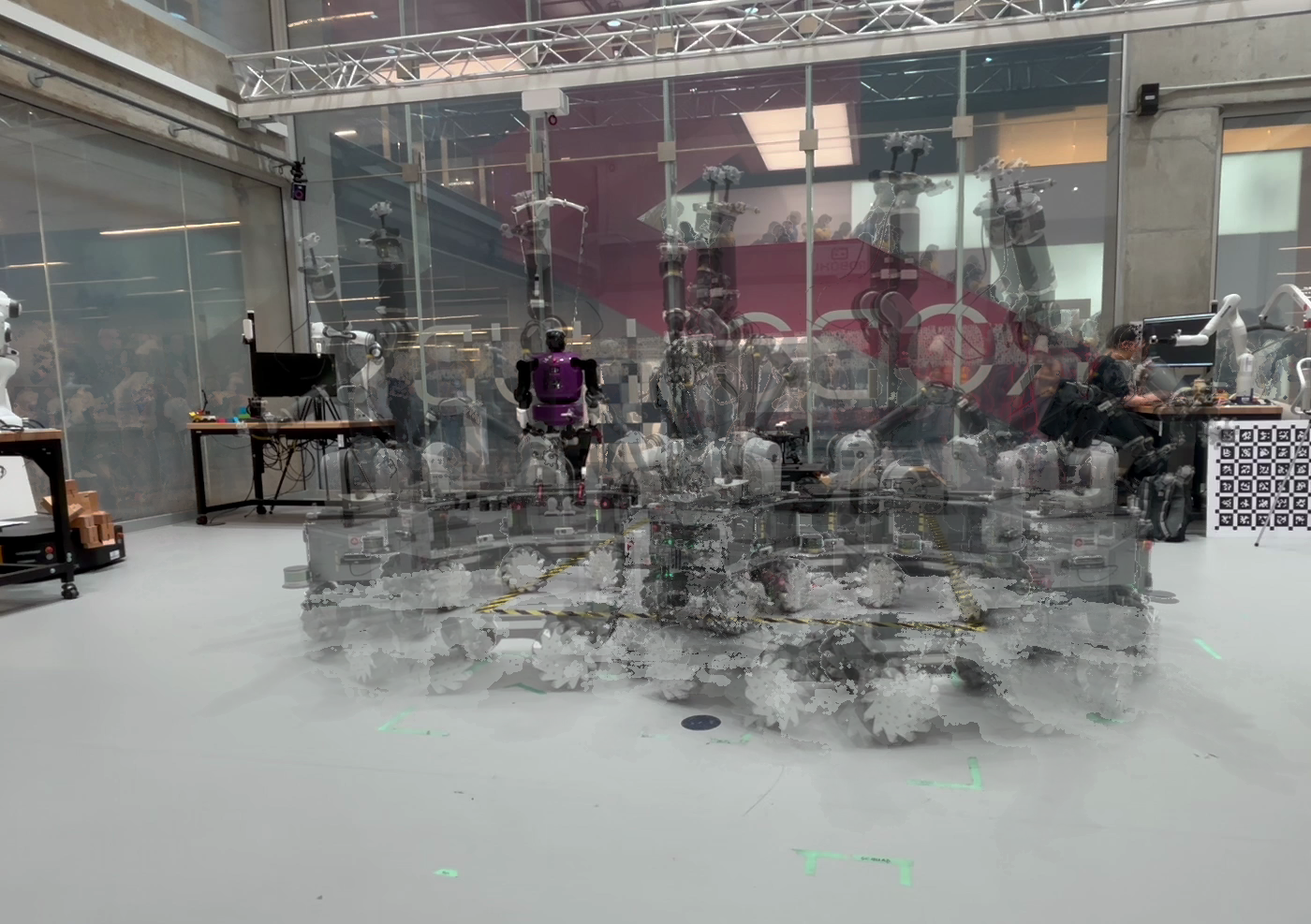}}
    \\
    \caption{A collection of box plots, 3D trajectory plots, and time-lapsed images for dual-\gls{VIO} (without loop-closure) maneuvering in seven different base motions with arm sweeping up and down (UD). The box plots (a,b,c,d) demonstrate the \acrshort{ATE} in terms of translation and orientation components for the base and \gls{EE} \gls{VIO} modules, respectively. The 3D trajectory plots, (e,f,g,h,i,j,k) and (l,m,n,o,p,q,r) illustrate the trajectories for the base and \gls{EE}, correspondingly. There are three trajectories in each plot; the black dashed line indicates \gls{GT} from \gls{Vicon}, the blue dash-dotted line indicates the baseline with two \gls{VINS-F} in parallel, and the solid orange line showcases our proposed method with tightly-coupled dual-\gls{VIO} methods through arm odometry. The seven base maneuvers at the holonomic base are indicated by time-lapse images (s,t,u,v,w,x,y), generated by super-imposing the sampled frames per $2\units{s}$ from the recorded video. \label{fig:result:UD}}
\end{figure*}
\subsection{Performance Results \label{sec:perf-results}}
Among five arm motions, the up-down arm motions are the most intricate dynamic locomotion with seven different base maneuvers. In Fig. \ref{fig:result:UD}, the box plots \ref{fig:result:UD:base:P} and \ref{fig:result:UD:EE:P} summarize the translation error for both the base \gls{VIO} and the \gls{EE} \gls{VIO}. It is apparent that the translation error and its variance are significantly lower and more consistent for both \gls{VIO} modules at the base and especially at the \gls{EE}. However, the improvements in orientation are not as significant as shown in \ref{fig:result:UD:base:R} and \ref{fig:result:UD:EE:R}.

\begin{figure*}[t]
    \centering
    \vspace{-1em} 
    \subfloat[\gls{EE} translation \acrshort{ATE} metrics with arm pointing up (U)\label{fig:result:U:EE:P}]{\includegraphics[width=\columnwidth]{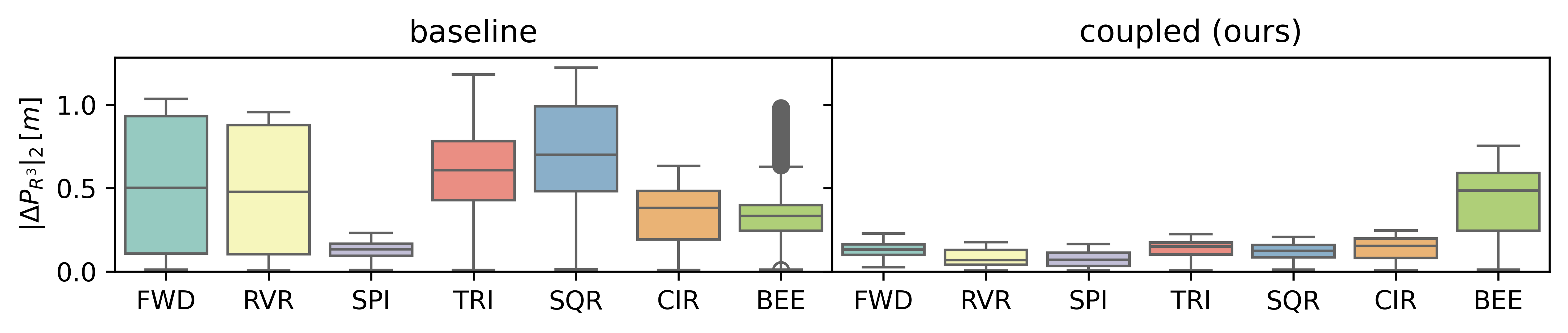}}\;
    \subfloat[Base translation \acrshort{ATE} metrics with arm pointing up (U) \label{fig:result:U:base:P}]{\includegraphics[width=\columnwidth]{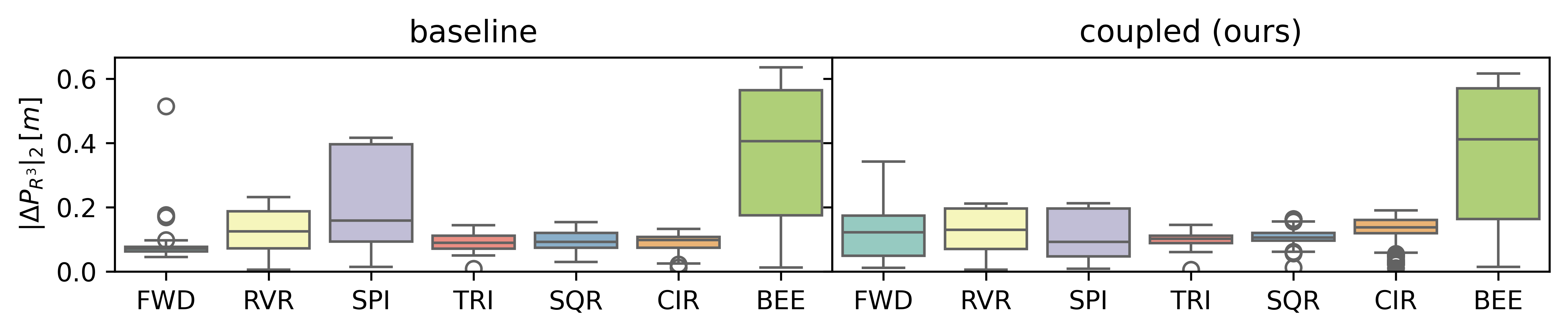}}
    \\
    \vspace{-1em} 
    \subfloat[\gls{EE} translation \acrshort{ATE} metrics with arm pointing down (D)\label{fig:result:D:EE:P}]{\includegraphics[width=\columnwidth]{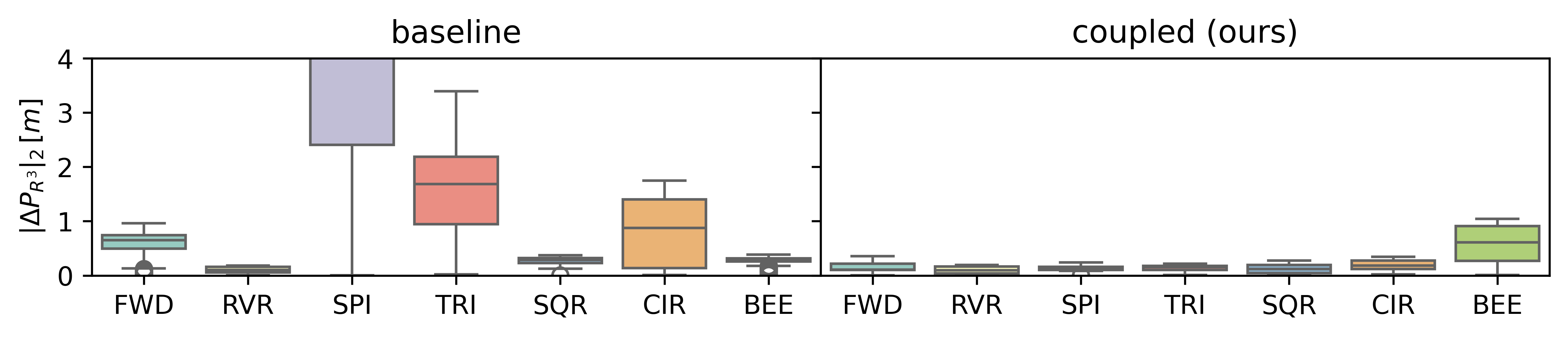}}\;
    \subfloat[Base translation \acrshort{ATE} metrics with arm pointing down (D) \label{fig:result:D:base:P}]{\includegraphics[width=\columnwidth]{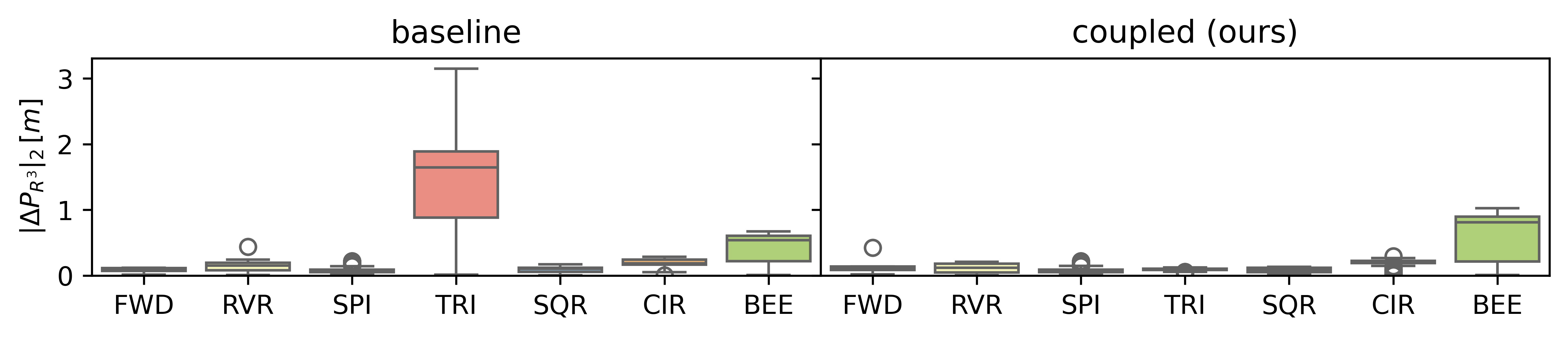}}
    \\
    \vspace{-1em} 
    \subfloat[\gls{EE} translation \acrshort{ATE} metrics with arm extending forward (E) \label{fig:result:E:EE:P}]{\includegraphics[width=\columnwidth]{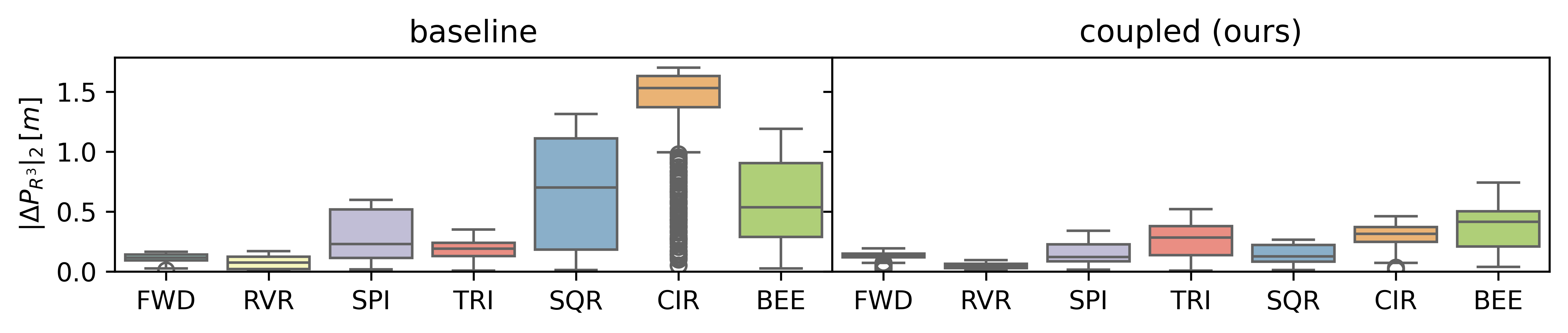}}\;
    \subfloat[Base translation \acrshort{ATE} metrics with arm extending forward (E) \label{fig:result:E:base:P}]{\includegraphics[width=\columnwidth]{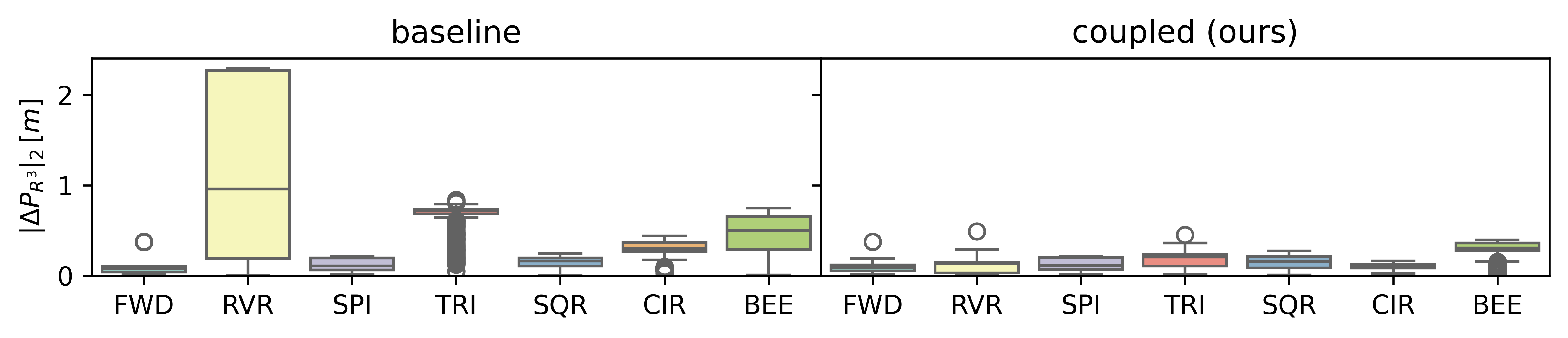}}
    \\
    \vspace{-1em} 
    \subfloat[\gls{EE} translation \acrshort{ATE} metrics with left-right arm \REVISE{motion} (LR)\label{fig:result:LR:EE:P}]{\includegraphics[width=\columnwidth]{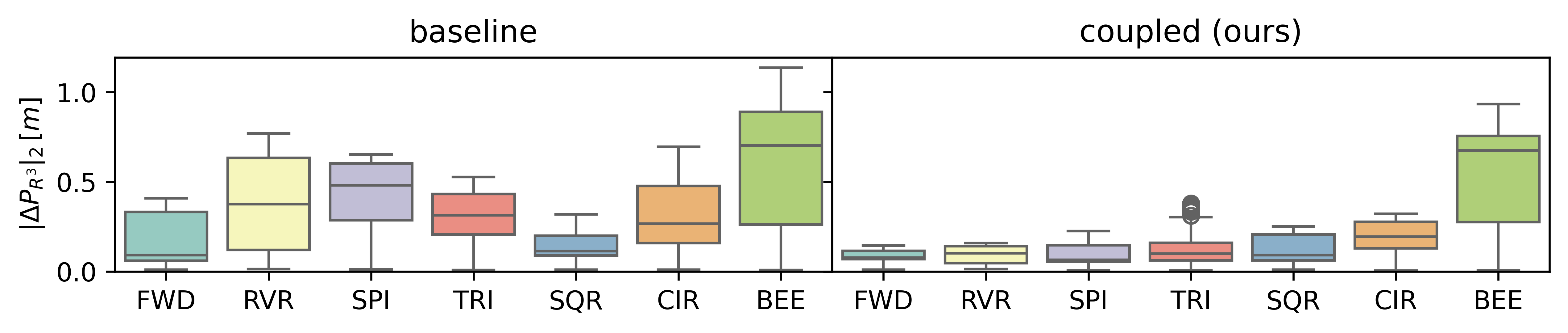}}\;
    \subfloat[Base translation \acrshort{ATE} metrics with left-right arm \REVISE{motion} (LR) \label{fig:result:LR:base:P}]{\includegraphics[width=\columnwidth]{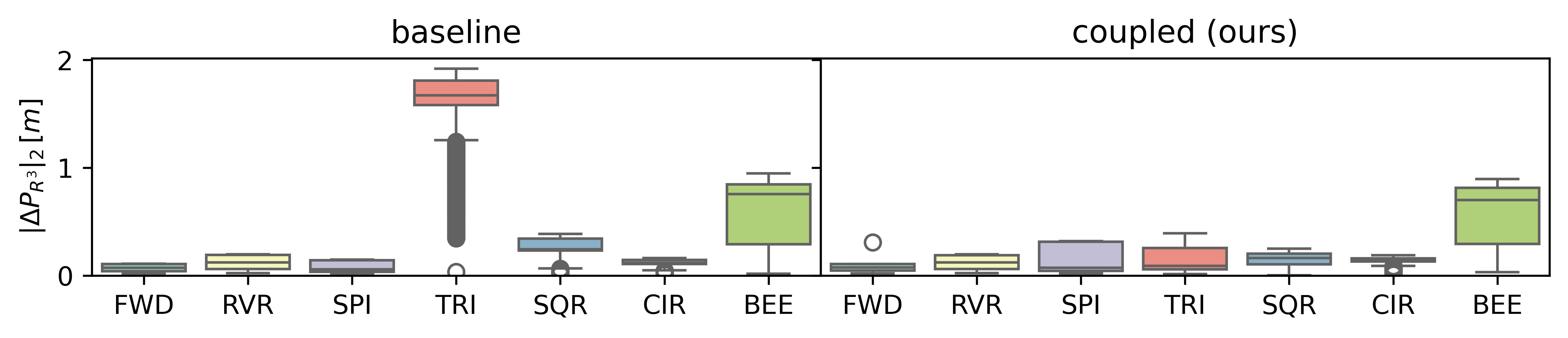}}
    \\
    \caption{A collection of box plots for dual-\gls{VIO} (without loop-closure) maneuvering in all seven different base motions with four different arm motions (U, D, E, LR). The box plots (a,c,e,g) and (b,d,f,h) summarize the \acrshort{ATE} in terms of translation components for the base and the \gls{EE} \glspl{VIO}, respectively.\label{fig:result:6x6}}
\end{figure*}

The rest of the locomotion (\ie{} four arm motions by seven base movements) is summarized in Fig. \ref{fig:result:6x6} with the translational component of \acrshort{ATE} for both \gls{VIO} modules at the \gls{EE} and the base. Similar to the box plots in Fig. \ref{fig:result:UD}, the results in Fig. \ref{fig:result:6x6} also suggest that our proposed method significantly improves the accuracy and consistency of the \gls{VIO} at the \gls{EE}, particularly when the locomotion becomes more complex. This should come as no surprise; the \gls{EE} suffers from significant vibrations and excitation when the mobile base crosses the uneven terrain due to the imperfect ground levelling in the testing facility and stiff omni-wheels. By imposing the base \gls{VIO} onto the \gls{EE} \gls{VIO} pose graph optimization, the resultant pose estimation has been made substantially better, thereby producing robust odometry with minimal consistent \gls{ATE}. 

Meanwhile, performance of the base odometry appears generally similar to that of the baseline; however, it showed more consistency throughout different base motions. This is because our methods can suppress significant odometry errors from the ground impacts or visual ambiguities through the regulation of the complementary \gls{VIO} estimates and thus can prevent early divergence of the \glspl{VIO} upon encountering uncertainties.

As shown in Table \ref{tab:result:ate}, the \REVISE{mean} \gls{ATE} \REVISE{in translation} is smaller and more consistent than the baseline data (as highlighted in bold) for the dynamic locomotion involving complex base maneuvers (SPI, TRI, SQR, CIR, BEE) and active arm motions (LR, UD). \REVISE{The table for orientation ATE is omitted here due to page limit, but their values for our method also showed improvements over baseline results, although not as drastic as the translation ATE in Table \ref{tab:result:ate}. (Also, see Figs. \ref{fig:result:UD:base:R} and \ref{fig:result:UD:EE:R}).}

\begin{table}[!h]
    \centering
    \caption{Mean \acrshort{ATE} in translation at \glspl{VIO} with dynamic locomotion}
    \label{tab:result:ate}
    \setlength{\tabcolsep}{3pt}
    \begin{tabular}{|l*{2}{|| c c | c c }|}
    \hline
    {} & \multicolumn{4}{c||}{VINS-Fusion (dual) $\units{m}$} & \multicolumn{4}{c|}{Ours (dual+odom) $\units{m}$} 
    \Bstrut\Tstrut
    \\ \cline{2-5} \cline{6-9}
    {} & \multicolumn{2}{c|}{Left-Right} & \multicolumn{2}{c||}{Up-Down} & \multicolumn{2}{c|}{Left-Right} &  \multicolumn{2}{c|}{Up-Down}\Tstrut
    \\ 
    {} & EE & Base & EE & Base & EE & Base & EE & Base 
    \Bstrut\Tstrut\\ 
    \hline
    FWD & 0.179 &     {0.071} {}&     { 1.966} &    {0.076} {}& \bf{0.089} &     0.075  {}& \bf{0.188} &     0.086\Tstrut\\ 
    RVR & 0.389 &      0.120  {}&       0.277  &    {0.073} {}& \bf{0.094} & \bf{0.119} {}& \bf{0.264} &     0.110\Bstrut\\ 
    \hline
    SPI & 0.447 &     {0.076} {}&     { 4.884} &     0.196  {}& \bf{0.094} &     0.146  {}& \bf{0.173} & \bf{0.121}\Tstrut\\
    TRI & 0.309 &     {1.582} {}&       0.402  &    {0.146} {}& \bf{0.142} & \bf{0.149} {}& \bf{0.121} &    {0.150}\\ 
    SQR & 0.147 &      0.264  {}&     {45.691} &     0.134  {}& \bf{0.125} & \bf{0.150} {}& \bf{0.168} & \bf{0.127}\\
    CIR & 0.324 &     {0.115} {}&       0.560  &     0.147  {}& \bf{0.197} &     0.134  {}& \bf{0.137} & \bf{0.132}\\
    BEE & 0.586 &      0.579  {}&       0.502  &     0.347  {}& \bf{0.537} & \bf{0.554} {}& \bf{0.277} & \bf{0.199}\Bstrut\\ 
    \hline        
    \end{tabular}
\end{table}

\REVISE{Note that our method outperforms the baseline \gls{VIO} not only at the \gls{EE} but also at the base. Expectedly, the benefit of tight coupling between the \gls{EE} \gls{VIO} and the base \gls{VIO} is more pronounced at the \gls{EE} because the base \gls{VIO} is generally more stable than the \gls{EE} \gls{VIO}, and thus fusing them offers major benefits to enhancing the accuracy of the \gls{EE} \gls{VIO}. However, the base \gls{VIO} of our method also outperforms the baseline data in some cases. This is because the estimation from the \gls{EE} \gls{VIO} can mitigate some errors at the base \gls{VIO} caused by uneven terrains and dynamic coupling with the arm motion.} 

\section{Conclusions and Future Work \label{sec:conclusion}}

A new tightly coupled dual-\gls{VIO} scheme has been proposed for robust dual-state estimation of a mobile manipulator operating under dynamic locomotion. It moderates arm joint angles for both \gls{VIO} modules (the mobile base and the \gls{EE}) through arm kinematics. This fusion enables handling of degenerate modes through arbitration, thereby bringing the best out of each system; stable odometry from the base and dexterous \gls{FOV} from the \gls{EE}. 
Our approach adopts jointly optimizing two \gls{VIO} modules without maintaining a single large factor graph, allowing real-time capability for multi-\gls{VIO} fusion for kinematically constrained systems with aggressive dynamic locomotion. This work may be useful for active-\gls{SLAM}, where the \gls{EE} \gls{VIO} can focus on active exploration for mapping, while the base \gls{VIO} concentrates on exploitation for re-localization.

The current implementation has three major uncertainties in the arm; unobservable modes in modelling, inaccurate zeroing, and inaccurate eye-in-hand pose between the camera and the robot. These factors limit the accurate estimation of both orientation and translation for basic motions. Therefore, there is a need for an online calibration to improve the performance further. \REVISE{As another future work, the proposed dual-VIO may be used in stereo mode to further improve the accuracy with the help of high bandwidth communication channel. Furthermore, we also plan to utilize mutual features between two cameras to support online temporal calibration on the fly, realizing fault tolerance and self-reinitialization capability.}


\bibliography{3_Reference_Survey-2023-Nov-1-Dec-21.bib}
\begin{IEEEbiography}[{\includegraphics[width=1in,height=1.25in,clip,keepaspectratio]{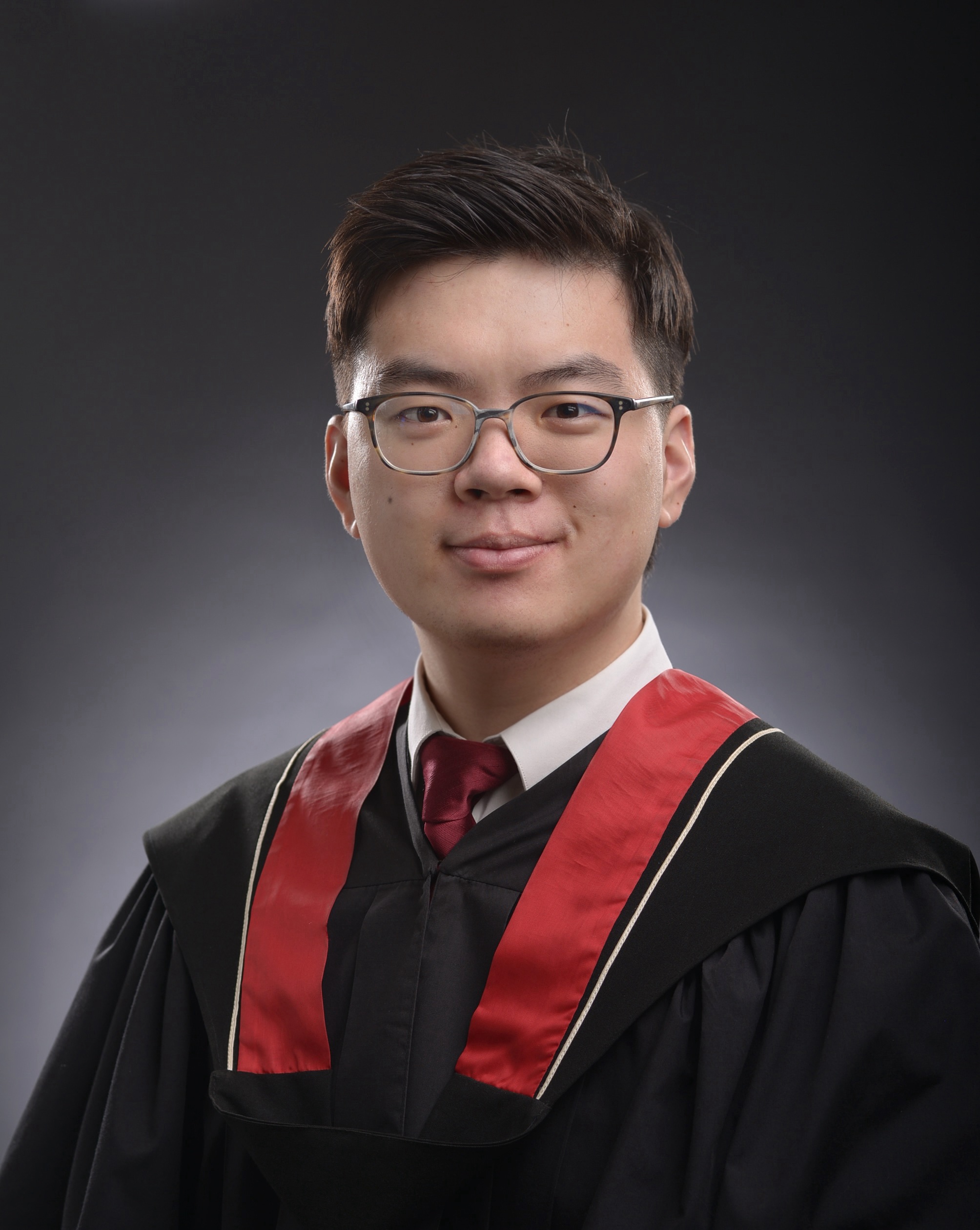}}]{Jianxiang (Jack) Xu} received the B.S. degree at University of Waterloo, with distinction in Mechatronics Engineering, Honours, Co-operative Program with Artificial Intelligence Option in 2021. He is pursuing an M.Sc. degree in Mechanical and Mechatronics Engineering at the University of Waterloo and researching at the Waterloo Mechanical Systems \& Control Lab. His research interests include robotics, mechatronics, system architecture, controls, state estimation, computer vision, and machine learning. He is a student member of ASME 2024.
\end{IEEEbiography}
\begin{IEEEbiography}[{\includegraphics[width=1in,height=1.25in,clip,keepaspectratio]{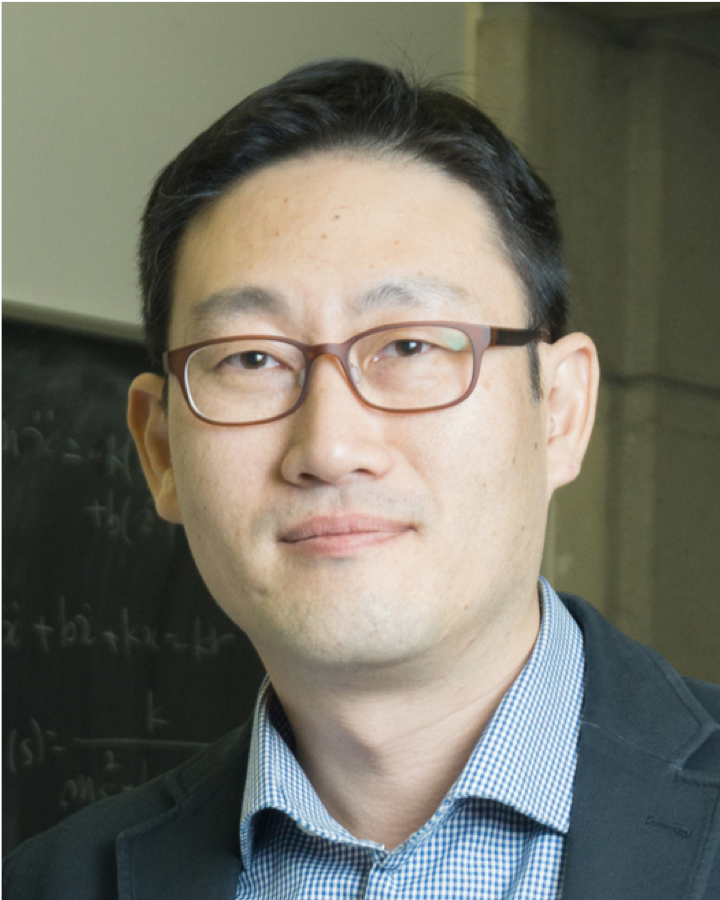}}]{Soo Jeon}
received his B.S. and M.S. degrees from Mechanical \& Aerospace Engineering at Seoul National University, Korea in 1998 and 2001 respectively, and his Ph.D. from Mechanical Engineering at the University of California, Berkeley in 2007. After graduation, he worked as a mechanical engineer in Applied Materials Inc. until he moved to the Department of Mechanical \& Mechatronics Engineering at the University of Waterloo in 2009, where he is currently a full professor. His research interests include dynamic systems and control, mechatronic system design, friction-induced stability and machine learning for physical systems. His research applications cover robotics, industry automation, medical ultrasound, and transportation systems. He received Rudolf Kalman Best Paper Award from ASME Dynamic Systems and Control Division (DSCD) in 2010, the Discovery Accelerator Supplement Award from NSERC (Natural Sciences and Engineering Research Council) of Canada in 2015, the Best Paper Award in Robotics at CRV (Conference on Robots and Vision) in 2022, and the Engineer of the Year Award from KOFST in 2022. He is a member of ASME, IEEE, CSME (Canadian Society for Mechanical Engineering) and PEO (Professional Engineers Ontario). He has been an associate editor for the ASME Journal of Dynamic Systems, Measurement and Control, IEEE Transactions on Automation Science and Engineering, IEEE/ASME Transactions on Mechatronics, and IEEE Robotics and Automation Letters.
\end{IEEEbiography}
\vfill
\end{document}